\def\year{2018}\relax
\documentclass[letterpaper]{article} 
\usepackage{aaai18_dummy}  
\usepackage{times}  
\usepackage{helvet}  
\usepackage{courier}  
\usepackage{url}  
\usepackage{graphicx}  
\usepackage{algorithm}
\usepackage{algorithmic}
\usepackage{amsfonts}
\usepackage{amsmath}
\usepackage{subfigure}
\newcommand{\vect}[1]{{\boldsymbol{#1}}}

\newcommand{\Real}{\mathbb{R} }
\newcommand{\E}{\mathbb{E} }

\begin{document}
	%
	\title{Hierarchical Policy Search via Return-Weighted Density Estimation}
	\author{Takayuki Osa\\
		University of Tokyo\\
		277-0882, Chiba, Japan\\
		RIKEN Center for AIP\\
		103-0027, Tokyo, Japan\\
		\And Masashi Sugiyama \\
		RIKEN Center for AIP\\
		103-0027, Tokyo, Japan\\
		University of Tokyo\\
		277-0882, Chiba, Japan\\
	}
	\maketitle
	\begin{abstract}
		Learning an optimal policy from a multi-modal reward function
		is a challenging problem in reinforcement learning (RL).
		\emph{Hierarchical} RL
		(HRL) tackles this problem by learning a hierarchical policy,
		where multiple option policies are in charge of
		different strategies corresponding to modes of a reward function
		and a gating policy selects the best option for a given context.
		Although HRL has been demonstrated to be promising,
		current state-of-the-art methods cannot still perform well in complex real-world problems
		due to the difficulty of identifying modes of the reward function.
		In this paper, we propose a novel method called
		\emph{hierarchical policy search via return-weighted density estimation} (HPSDE),
		which can efficiently identify the modes
		through density estimation with return-weighted importance sampling.
		Our proposed method finds option policies corresponding to the modes of the return function
		and automatically determines the number and the location of option policies,
		which significantly reduces the burden of hyper-parameters tuning.
		Through experiments, we demonstrate that the proposed HPSDE
		successfully learns option policies corresponding to modes of the return function and that it
		can be successfully applied to a challenging motion planning problem
		of a redundant robotic manipulator.
	\end{abstract}

\section{Introduction}
Recent work on reinforcement learning~(RL) has been successful in various tasks, including robotic manipulation~\cite{Gu17,Levine16,Levine16b} and playing a board game~\cite{silver16}.
However, many RL methods cannot leverage a hierarchical task structure,
whereas many tasks in the real world are highly structured.
Grasping is a good example of such structured tasks.
When grasping an object, humans know multiple grasp types from their experience and 
adaptively decide how to grasp the given object~\cite{Cutkosky90,Napier56}.
This strategy can be interpreted as a hierarchical policy where the gating policy first selects the grasp type 
and the option policy that represents the selected grasp type subsequently plans the grasping motion~\cite{Osa16}. 
Prior work on hierarchical RL suggests that learning various option policies increases the versatility~\cite{Daniel16}
and that exploiting a hierarchical task structure can exponentially reduce the search space~\cite{Dietterich00}. 

However, RL of a hierarchical policy is not a trivial problem.
As indicated by \citeauthor{Daniel16}~(\citeyear{Daniel16}), each option policy in a hierarchical policy needs to focus on a single mode of the return function, 
otherwise a learned policy will try to average multiple modes and fall into a local optimum with poor performance.
Therefore, it is necessary to properly assign option policies to individual modes of the return function,
and the challenge of this problem is how to identify the number and locations of modes of the return function.
Although regularizers~\cite{Bacon17,Florensa17} can be used to drive the option policies to various solutions,
they cannot prevent an option policy from averaging over multiple modes of the return function.
Additionally, in existing methods, the performance often depends on initialization or pre-training of the policy~\cite{Daniel16,Florensa17},
and a user often needs to specify the number of the option policies in advance, which significantly affects the performance~\cite{Bacon17}. 

To address such issues in existing hierarchical RL, 
we propose model-free hierarchical policy search via return-weighted density estimation (HPSDE).
Our approach reduces the problem of identifying the modes of the return function 
to estimating the return-weighted sample density.
Unlike previous methods, the number and the location of the option policies is automatically determined without explicitly estimating the option policy parameters, and an option policy learned by HPSDE focuses on a single mode of the return function.
We discuss the connection between the expected return maximization and the density estimation with return-weighted importance.
The experimental results show that HPSDE outperforms a state-of-the-art hierarchical RL method
and that HPSDE finds multiple solutions in motion planning for a robotic redundant manipulator. 

\section{Problem Formulation}
\label{sec:find_modes}
We consider a reinforcement learning problem in the Markov decision process (MDP) framework, where an agent is in a state $\vect{x} \in \mathcal{X}$ and takes an action $\vect{u} \in \mathcal{U}$.
In this paper, we concentrate on the episodic case of hierarchical RL, where a policy $\pi( \vect{\tau} | \vect{x}_{0} )$ generates a trajectory $\vect{\tau}$ for a given initial state $\vect{x}_{0}$ and only one selected policy is executed until the end of the episode.
After every episode, the agent receives a return of a trajectory $R(\vect{\tau}, \vect{x}_{0}) = \sum^{T}_{t=0} r_{t} $, which is given by a sum of immediate rewards. $T$ is the length of the trajectory, which is a random variable.
In the following, we denote by $\vect{s} = \vect{x}_{0}$ the initial state of the episode, which is often referred to as the ``context'', 
and we assume that a trajectory $\vect{\tau}$ contains all the information of actions $\vect{u}_{t}$ and the state $\vect{x}_{t}$ during the episode.
The purpose of policy search is to obtain the policy $\pi( \vect{\tau} | \vect{s} )$ that maximizes the expected return~\cite{Deisenroth13} 
\begin{align}
J(\pi) 
= \iint d(\vect{s}) \pi( \vect{\tau} | \vect{s} ) R(  \vect{s}, \vect{\tau}  )  \textrm{d}\vect{\tau} \textrm{d}\vect{s},
\end{align}
where $d(\vect{s} )$ is the distribution of the context $\vect{s}$.
In hierarchical RL, we consider a policy that is given by a mixture of option policies:
\begin{align}
\pi(\vect{\tau} | \vect{s}) = \sum_{o  \in \mathcal{O}} \pi(o | \vect{s})\pi(\vect{\tau} | \vect{s}, o),
\end{align}
where $o$ is a latent variable which represents the option of the policy and $\mathcal{O}$ is a set of the latent variable $o$ in a hierarchical policy. We refer to $\pi(\vect{\tau} | \vect{s}, o)$ as an option policy and $\pi(o | \vect{s})$ as a gating policy.
Here, the option is defined over trajectories, unlike the option over actions as in \citeauthor{Machado17}~(\citeyear{Machado17}); \citeauthor{Bacon17}~(\citeyear{Bacon17}).
The expected return of a hierarchical policy is given by
\begin{align}
J(\pi) 
= \sum_{o  \in \mathcal{O}} { \iint d(\vect{s}) \pi( o | \vect{s} ) \pi( \vect{\tau} | o, \vect{s} ) R( \vect{s}, \vect{\tau} ) \textrm{d}\vect{\tau}\textrm{d}\vect{s} }.
\label{eq:return_HRL}
\end{align}
The goal of hierarchical policy search is to learn an optimal hierarchical policy $\pi^{*}$  that maximizes the expected return $J(\pi)$:
\begin{align}
\pi^{*}(\vect{\tau} | \vect{s}) = \sum_{o \in \mathcal{O} } \pi^{*}(o | \vect{s})\pi^{*}(\vect{\tau} | \vect{s}, o) = \arg \max_{\pi} J(\pi).
\label{eq:objective}
\end{align}
This problem can be solved with respect to samples $\mathcal{D} = \{\left( \vect{s}_{i}, \vect{\tau}_{i}, R_{i}(\vect{s}_{i}, \vect{\tau}_{i}) \right) \}$
collected through rollouts. 
The challenge of hierarchical policy search is to estimate the latent variable $o$, which we cannot observe.
The expected return for the policy $\pi$ in (\ref{eq:return_HRL}) can be rewritten as 
\begin{align}
J(\pi) 
= \sum_{o \in \mathcal{O} } p(o) \E[ R | \pi(\vect{\tau} | \vect{s}, o),  o ],
\label{eq:hps}
\end{align}
where the expectation $\E[ R | \pi(\vect{\tau} | \vect{s}, o),  o ]$ is given by
\begin{align}
\E[ R | \pi(\vect{\tau} | \vect{s}, o),  o ] =  \iint d(\vect{s} | o ) \pi(\vect{\tau} | \vect{s}, o) R(\vect{s}, \vect{\tau}) \textrm{d}\vect{s} \textrm{d}\vect{\tau},
\end{align}
and $d(\vect{s} | o )$ is the distribution of contexts $\vect{s}$ assigned to the option $o$.
When the assignment of each sample $(\vect{s}_{i}, \vect{\tau}_{i}, R(\vect{s}_{i}, \vect{\tau}_{i}) )$ to the option $o$ is known, 
maximizing $\E[ R | \pi(\vect{\tau} | \vect{s}, o),  o ]$ is equivalent to solving a policy search problem with respect to samples $\{ ( \vect{s}^{o}_{i}, \vect{\tau}^{o}_{i}, R(\vect{s}^{o}_{i}, \vect{\tau}^{o}_{i}) ) ) \}$  assigned to the option $o$.
Therefore, the formulation in (\ref{eq:hps}) indicates that, if we can estimate the latent variable $o_{i}$ for each sample $(\vect{s}_{i}, \vect{\tau}_{i})$, the option policy $\pi(\vect{\tau} | \vect{s}, o)$ can be learned using a policy search method for a monolithic policy, e.g., trust region policy optimization~(TRPO)~\cite{Schulman15} or relative entropy policy search~(REPS)~\cite{Peters10}.

As discussed above, estimation of the latent variable $o$ is a crucial problem in hierarchical RL. 
Next, we introduce the return-weighted density estimation to address this problem.

\section{Hierarchical Policy Search via Return-Weighted Density Estimation}
In this section, we introduce our proposed method, hierarchical policy search via return-weighted density estimation~(HPSDE).
To learn the number and the location of option policies, we estimate the latent variable $o$ through estimating the trajectory density induced by the unknown optimal policy $\pi^{*}$.
This density estimation is performed by using the return-weighted importance sampling.
In the following, we discuss the details of 1)~the return-weighted importance weight, 2)~the latent variable estimation, and 3)~the gating policy for selecting the option policies. 

\subsection{Return-Weighted Density Estimation}

To estimate the latent variable $o$, we consider the trajectory density induced by the optimal policy, 
$p^{*}( \vect{s}, \vect{\tau} ) = d(\vect{s} )\pi^{*}(\vect{\tau} | \vect{s})$.
Here, we assume that the optimal policy draws samples that lead to higher returns with higher probability.
This assumption is equivalent to assuming that the optimal policy $\pi^{*}( \vect{\tau} | \vect{s} )$ is of the form
\begin{equation}
\pi^{*}( \vect{\tau} | \vect{s} ) = \frac{ f \left( R(\vect{s}, \vect{\tau}) \right)}{ Z },
\label{eq:optimal_policy}
\end{equation}
where $Z$ is a partition function, and $f(R)$ is a \textit{functional}, which is a function of the return function.
$f(R)$ should be monotonically increasing with respect to $R$ such that a trajectory with a higher return is generated with higher probability by the optimal policy.  
This assumption is commonly used in prior work on policy search~\cite{Deisenroth13}. 
Under this assumption, finding the modes of the return function $R( \vect{s}, \vect{\tau} )$ is equivalent to finding the modes of the density $p^{*}( \vect{s}, \vect{\tau} )$ induced by the optimal policy $\pi^{*}( \vect{\tau} | \vect{s} )$.

Since the optimal policy $\pi^{*}( \vect{\tau} | \vect{s} )$ is unknown,
we cannot directly sample $\{ (\vect{s}^{*}_{i}, \vect{\tau}^{*}_{i}) \} \sim d(\vect{s})\pi^{*}( \vect{\tau} | \vect{s} )$.
Thus, we use an importance sampling technique to estimate the density $d(\vect{s})\pi^{*}( \vect{\tau} | \vect{s} )$ induced by the optimal policy $\pi^{*}( \vect{\tau} | \vect{s} )$.

We collect samples $\{(  \vect{s}_{i}, \vect{\tau}_{i}, R(\vect{s}_{i}, \vect{\tau}_{i}) ) \}^{N}_{i=1}$ drawn from the current policy $\pi_{\textrm{old}}( \vect{\tau} | \vect{s} )$ and a given context distribution $d(\vect{s})$.
The importance of the sample $(  \vect{s}_{i}, \vect{\tau}_{i})$ can be given by
\begin{align}
W( \vect{s}_{i}, \vect{\tau}_{i} )
& = \frac{d(\vect{s}_{i})\pi^{*}(\vect{\tau}_{i} | \vect{s}_{i} )}{d(\vect{s}_{i})\pi_{\textrm{old}}(\vect{\tau}_{i} | \vect{s}_{i})} 
= \frac{\pi^{*}(\vect{\tau}_{i} | \vect{s}_{i} )}{\pi_{\textrm{old}}(\vect{\tau}_{i} | \vect{s}_{i})}  \\
& = \frac{  f \left( R(\vect{s}_{i}, \vect{\tau}_{i} ) \right) }{ Z \pi_{\textrm{old}}(\vect{\tau}_{i} | \vect{s}_{i})}.
\end{align}
Note that the context distribution $d(\vect{s})$ is given by an environment and invariant to the policy in an episodic case.
If we normalize $W$, we obtain the normalized weight
\begin{align}
&\tilde{W}( \vect{s}_{i}, \vect{\tau}_{i}) 
= \frac{W(\vect{s}_{i}, \vect{\tau}_{i}) }{ \sum^{N}_{j=1} W( \vect{s}_{j}, \vect{\tau}_{j} )  } \\
& = \frac{ \frac{  f \left( R(\vect{s}_{i}, \vect{\tau}_{i} ) \right) }{ Z \pi_{\textrm{old}}(\vect{\tau}_{i} | \vect{s}_{i})} }{ \sum^{N}_{j=1} \frac{  f \left( R(\vect{s}_{j}, \vect{\tau}_{j} ) \right) }{ Z \pi_{\textrm{old}}(\vect{\tau}_{j} | \vect{s}_{j})} } 
= \frac{ \frac{  f \left( R(\vect{s}_{i}, \vect{\tau}_{i} ) \right) }{ \pi_{\textrm{old}}(\vect{\tau}_{i} | \vect{s}_{i})} }{ \sum^{N}_{j=1} \frac{  f \left( R(\vect{s}_{j}, \vect{\tau}_{j} ) \right) }{ \pi_{\textrm{old}}(\vect{\tau}_{j} | \vect{s}_{j})} }.
\label{eq:normalized_w}
\end{align}
Therefore, we can compute the importance $\tilde{W}( \vect{s}_{i}, \vect{\tau}_{i})$  of each sample even though the partition function $Z$ of the optimal policy $\pi^{*}$ is unknown. 
Thus, we can estimate the density induced by the optimal policy $d(\vect{s})\pi^{*}( \vect{\tau} | \vect{s} )$ by using $\{(  \vect{s}_{i}, \vect{\tau}_{i}, R(\vect{s}_{i}, \vect{\tau}_{i}) ) \}^{N}_{i=1} \sim d(\vect{s})\pi_{\textrm{old}}( \vect{\tau} | \vect{s} )$ with the importance weight $\tilde{W}$.
We refer to the density estimation using the importance weight $\tilde{W}$ as \textit{return-weighted density estimation}.

Learning the optimal policy can be formulated as the distribution matching between the trajectory densities induced by the optimal policy and the learner's policy
\begin{align}
\hat{\pi}(\vect{\tau} | \vect{s}) = \arg \min_{\pi} D_{\textrm{KL}}( p^{*}(\vect{s}, \vect{\tau}) || p_{\pi}(\vect{s}, \vect{\tau} ) ),
\end{align}
where $ D_{\textrm{KL}}( p^{*}(\vect{s}, \vect{\tau}) || p_{_{\pi}}(\vect{s}, \vect{\tau} ) )$ is the KL divergence  given by
\begin{align}
& D_{\textrm{KL}}( p^{*}(\vect{s}, \vect{\tau}) ||p_{_{\pi}}(\vect{s}, \vect{\tau} ) ) 
= \iint p^{*}(\vect{s, \vect{\tau}})\log \frac{p^{*}(\vect{s, \vect{\tau}})}{ p_{\pi}(\vect{s}, \vect{\tau} )} \textrm{d}\vect{\tau}\textrm{d}\vect{s} \nonumber \\
& = \iint W(\vect{s, \vect{\tau}}) p_{\pi_{\textrm{old}}}(\vect{s, \vect{\tau}})\log \frac{ W(\vect{s, \vect{\tau}}) p_{\pi_{\textrm{old}}}(\vect{s, \vect{\tau}})}{ p_{\pi}(\vect{s}, \vect{\tau} )} \textrm{d}\vect{\tau}\textrm{d}\vect{s}, 
\label{eq:kl_p}
\end{align}
$p^{*}(\vect{s}, \vect{\tau})$ is the density induced by the optimal policy, $p_{\pi}(\vect{s}, \vect{\tau} )$ is the density induced by the newly learned policy, and $p_{\pi_{\textrm{old}}}(\vect{s, \vect{\tau}})$ is the density induced by the old policy used for data collection.
The minimizer of $ D_{\textrm{KL}}( p^{*}(\vect{s}, \vect{\tau}) || p_{\pi}(\vect{s}, \vect{\tau} ) )$ is given by the maximizer of the weighted log likelihood:
\begin{align}
L( \pi, \pi_{\textrm{old}} ) & = \iint  W(\vect{s, \vect{\tau}}) p_{\pi_{\textrm{old}}}(\vect{s, \vect{\tau}}) \log p_{\pi}(\vect{s}, \vect{\tau} ) \textrm{d}\vect{\tau}\textrm{d}\vect{s} \nonumber \\
& \approx \frac{1}{N} \sum_{ (\vect{s}_{i}, \vect{\tau}_{i}) \in \mathcal{D} } \tilde{W}(\vect{s}_{i}, \vect{\tau}_{i}) \log p_{\pi}(\vect{s}_{i}, \vect{\tau}_{i} ).
\end{align}



The connection to maximizing the expected return can be seen as follows.
If we assume that $R > 0$ and the optimal policy follows $\pi^{*}( \vect{\tau} | \vect{s} ) = R(\vect{s}, \vect{\tau}) / Z $, in a manner similar to the results shown by \citeauthor{Dayan97}~(\citeyear{Dayan97}) and \citeauthor{Kober11}~(\citeyear{Kober11}), we can obtain 
\begin{align}
&\log J( \pi  ) 
= \log \iint p_{\pi}( \vect{s}, \vect{\tau} ) R( \vect{\tau}, \vect{s} ) \textrm{d}\vect{\tau}\textrm{d}\vect{s}
\nonumber \\
&= \log \iint p_{\pi_{\textrm{old}}}( \vect{s}, \vect{\tau} ) \frac{ d(\vect{s}) R( \vect{\tau}, \vect{s} )}{ p_{\pi_{\textrm{old}}}( \vect{s}, \vect{\tau} ) } 
\frac{p_{\pi}( \vect{s}, \vect{\tau} )}{d(\vect{s})} \textrm{d}\vect{\tau}\textrm{d}\vect{s} \nonumber \\
&\geq  \iint p_{\pi_{\textrm{old}}}( \vect{s}, \vect{\tau}) \frac{ d(\vect{s}) R( \vect{\tau}, \vect{s} )}{ p_{\pi_{\textrm{old}}}( \vect{s}, \vect{\tau} )  } 
\log \frac{p_{\pi}( \vect{s}, \vect{\tau} )}{d(\vect{s})} \textrm{d}\vect{\tau}\textrm{d}\vect{s} \nonumber \\
&=   Z \iint p_{\pi_{\textrm{old}}}( \vect{s}, \vect{\tau} ) \frac{ R( \vect{\tau}, \vect{s} )}{ \pi_{\textrm{old}}( \vect{\tau} | \vect{s} ) Z} 
\log  \frac{p_{\pi}( \vect{s}, \vect{\tau})}{d(\vect{s})} \textrm{d}\vect{\tau}\textrm{d}\vect{s} \nonumber \\
&\propto L( \pi, \pi_{\textrm{old}} ) + \textrm{const} .
\end{align}
Therefore, maximizing the weighted log likelihood $L( \pi, \pi_{\textrm{old}} )$ is equivalent to maximizing the lower bound of the expected return in this case.


In practice, the return function $R$ and $f(\cdot)$ in (\ref{eq:optimal_policy}) need to be designed by a practitioner.
To argue the optimal form of $f(\cdot)$ in (\ref{eq:optimal_policy}), we need to make further assumptions.
For example,  the optimal policy of the form $\pi^{*}( \vect{\tau} | \vect{s} ) = \exp \left( R(\vect{s}, \vect{\tau} )  \right)/ Z $ can be justified based on
the maximum entropy principle, which is often used in the literature of inverse RL~\cite{Ziebart10}.
In this work, we leave the form of $f(\cdot)$ in (\ref{eq:optimal_policy}) as an open design choice for a practitioner.




\subsection{Latent Variable Estimation}

Based on the discussion in the previous section, we estimate the latent variable $o$ of a hierarchical policy 
by matching the trajectory densities induced by the optimal policy and the learner's policy.
Specifically, we consider the likelihood:
\begin{align}
p( \vect{s}, \vect{\tau} )  = \sum_{o \in \mathcal{O}} p( o ) p( \vect{s}, \vect{\tau} | o )
\end{align}
and estimate the latent variable $o$ through approximating the trajectory density $p^{*}(\vect{s}, \vect{\tau})$ induced by the unknown optimal policy 
from samples $\{(  \vect{s}_{i}, \vect{\tau}_{i}, R(\vect{s}_{i}, \vect{\tau}_{i}) ) \}^{N}_{i=1}$ drawn from the current policy $\pi_{\textrm{old}}( \vect{\tau} | \vect{s} )$ using the return-weighted importance $\tilde{W}(\vect{s}_{i}, \vect{\tau}_{i})$.

To efficiently solve the estimation problem, we assume that an option policy is given by a Gaussian policy $\pi(\vect{\tau} | \vect{s}, o ) \sim \mathcal{N} \left( \vect{f}(\vect{s}), \vect{\Sigma} \right)$, which is frequently assumed in RL~\cite{Deisenroth13,Pirotta13,Furmston12}.
The mean $\vect{f}(\vect{s})$ can be linear to the feature vector as $\vect{w}^{\top}\vect{\phi}( \vect{s} )$ or represented by the output of a neural network.
If we assume that the region of the context $d(\vect{s} | o)$ for which the option policy is responsible is Gaussian,
$p( \vect{s}, \vect{\tau} | o ) = d(\vect{s} | o )\pi(\vect{\tau} | \vect{s}, o)$ should also be Gaussian.
Under these assumptions, the latent variable estimation is considered fitting a Gaussian mixture model.
Even if a cluster of samples $\{\vect{s}_{i}, \vect{\tau}_{i}\}$ is not Gaussian in practice, it will be represented by a mixture of Gaussian option policies.

The latent variable estimation for the Gaussian mixture model fitting can be performed by the expectation-maximization (EM) algorithm in general.
We can employ the variational Bayes expectation-maximization (VBEM) algorithm as well as the maximum likelihood~(ML) EM algorithm~\cite{Bishop06}.
The advantage of the VBEM algorithm over the ML EM algorithm is that the use of the symmetric Dirichlet distribution as a prior of the mixing coefficient leads to
a sparse solution.
This property of VBEM is preferable in our hierarchical policy search since it is likely that clusters found by VBEM focus on separate modes of the density and 
that unnecessary clusters are automatically eliminated.
Namely, we can obtain option policies corresponding to modes of the return function by using VBEM for the latent variable estimation. 

To obtain various option policies, \citeauthor{Daniel16}~(\citeyear{Daniel16})  employed a constraint to avoid the overlap between option policies and the number of the option policies was gradually reduced in the learning process.
Likewise, \citeauthor{Florensa17}~(\citeyear{Florensa17}) assumed that the number of the option policies is given by the user and modified the reward function based on mutual information in order to encourage visiting new states.
However, these constraints cannot prevent an option policy from averaging over multiple modes of the return function.
On the contrary, we do not employ such explicit constraints on option policies in HPSDE. 
By using a variational approach for estimating the latent variable, we can obtain a sparse solution which covers all the given samples with the minimum number of option policies.
Therefore, our approach can automatically determine the number of the option policies and each option policy obtained by HPSDE focuses on a separate single mode of the return function.


In practice, the dimensionality of $\vect{s}_{i}$ and $\vect{\tau}_{i}$ can be high and the shape of the cluster may not be Gaussian in the original feature space.
In such a case, nonlinear feature mapping with dimensionality reduction, e.g., the Laplacian eigenmaps~\cite{Belkin03}, can be used to perform the latent variable estimation properly.
In addition, trajectory representations such as Dynamic Movement Primitives (DMPs)~\cite{Ijspeert02} can also be used to represent a trajectory with a small number of parameters.

After estimating $p( o | \vect{s}_{i}, \vect{\tau}_{i} )$ for  each sample, 
each option policy $\pi(\vect{\tau} | \vect{s}, o)$ can be updated using an off-the-shelf policy search algorithm for learning a monolithic policy. 
In the next section, we describe the gating policy that selects the optimal option policy for a given context. 

\subsection{Selection of the Option Policy}
\label{sec:bandit}
When a context $\vect{s}$ is given by an environment and the option policies $ \pi(\vect{\tau} | \vect{s}, o)$ are learned,
the role of the gating policy $\pi(o | \vect{s})$ is to identify the option policy that maximizes the expected return for a given context.
Therefore, the gating policy is given by
\begin{align}
\pi(o | \vect{s}): o^{*} 
& = \arg \max_{o \in \mathcal{O}} \E\left[ R | \pi(\vect{\tau} | \vect{s}, o),  \vect{s}, o \right],
\end{align}
where the conditional expectation $\E\left[ R | \pi(\vect{\tau} | \vect{s}, o),  \vect{s}, o \right]$ is given by
\begin{align}
\E\left[ R | \pi(\vect{\tau} | \vect{s}, o),  \vect{s}, o \right]   = \int  \pi(\vect{\tau} | \vect{s}, o) R( \vect{s}, \vect{\tau}) \textrm{d}\vect{\tau}.
\label{eq:option_expectation}
\end{align}
Under the assumption that option policies are Gaussian, we explicitly estimate the expected return for each option policy $\E\left[ R | \pi(\vect{\tau} | \vect{s}, o), \vect{s}, o \right]$.
For this purpose, we approximate the return function with a Gaussian Process (GP)~\cite{Rasmussen05}
\begin{equation}
R( \vect{\tau}, \vect{s} ) \sim \mathcal{GP} \left( m \left( 
\boldsymbol{z}
\right), 
k \left( 
\vect{z},
\vect{z}'
\right) \right),
\label{eq:approximation}
\end{equation}
where $\vect{z} = [\vect{\tau}^{\top},  \vect{s}^{\top}]^{\top}$ and $m(\vect{z})$ is the mean.
For the kernel function $k\left( \vect{z}_{i}, \vect{z}_{j} \right)$, we employ the squared exponential kernel:
\begin{equation}
k \left( 
\vect{z}_{i},
\vect{z}_{j}
\right)
= \sigma^{2}_{f} \exp \left( - \frac{\lVert \boldsymbol{z}_{i} - \boldsymbol{z}_{j}  \rVert^{2} }{ 2l^{2} } \right) + \sigma_n^2 \delta_{\boldsymbol{z}_{i}\boldsymbol{z}_{j}},
\end{equation}
where $l$ is the bandwidth of the kernel, $\sigma^2_f$ is the function variance and $\sigma^2_n$ is the noise variance.
The hyperparameters $[l,\sigma_f, \sigma_n ]$ can be tuned by maximizing the marginal likelihood
using a gradient-based method~\cite{Rasmussen05}.

If we assume that the context is given and a trajectory is drawn from a Gaussian option policy $\pi( \vect{\tau} | \vect{s}, o ) \sim \mathcal{N}( \vect{\mu}^{o}(\vect{s}), \vect{\Sigma}^{o}(\vect{s}))$, $\vect{z}$ follows the Gaussian distribution:
\begin{align}
\vect{z} \sim \mathcal{N} \left( 
\vect{\mu}_{\vect{z}}, \vect{\Sigma}_{\vect{z}}
\right) ,
\end{align}
where 
\begin{align}
\vect{\mu}_{ \vect{z} } =
\left[
\begin{array}{c}
\vect{\mu}^{o}(\vect{s}) \\
\vect{s} 
\end{array}
\right], 
\vect{\Sigma}_{ \vect{z} } = 
\left[
\begin{array}{cc}
\vect{\Sigma}^{o}(\vect{s}) & 0\\
0 & 0
\end{array}
\right].
\end{align}
To compute the conditional expectation of the return given the option and the context in \eqref{eq:option_expectation},
we need to compute the following marginal distribution:
\begin{align}
p( R |  \vect{\mu}_{\vect{z}}, \vect{\Sigma}_{\vect{z}} ) = \int p( R | \vect{z}, \mathcal{D} )p( \vect{z} ) d\vect{z}.
\end{align}
The above marginal can be approximated with a Gaussian, and its mean $\E[R | \pi( \vect{\tau} | \vect{s}, o ), \vect{s}, o]$ and variance $\sigma_{R}$ can be analytically computed
~\cite{Girard02,Candela02}.

\begin{algorithm}[t]
	\caption{Hierarchical Policy Search via Return-Weighted Density Estimation (HPSDE) }
	\begin{algorithmic}
		\STATE{
			\textbf{Input:} the maximum number of the clusters $O_{\textrm{max}}$ \\
			Initialize the option policies, e.g. random policy \\
			Collect the rollout data $\mathcal{D} = \{ (\vect{s}_{i}, \vect{\tau}_{i}, R_{i}) \}$ by performing the initial policy
		}
		\REPEAT
		\STATE{ Compute the importance of each sample $\tilde{W}( \vect{s}_{i}, \vect{\tau}_{i})$\\
			Estimate $p( o | \vect{s}_{i}, \vect{\tau}_{i} )$ through density estimation using the importance weight $\tilde{W}$ \\
			Assign the samples $\{ (\vect{s}_{i}, \vect{\tau}_{i}) \}$ to option $o^{*}_{i} = \arg \max p( o | \vect{s}_{i}, \vect{\tau}_{i} )$\\
		}
		
		\FOR{ \textbf{each} $o$}
		\STATE{
			Train the $o$th policy using a policy search method\\ 
		}
		\ENDFOR \\
		\STATE{
			Train the GP model to approximate the return function\\
			Select the option $o^{*} = \arg \max \E\left[ R | \pi(\vect{\tau} | \vect{s}, o),  \vect{s}, o \right] $ \\ 
			Execute the rollout by following $\pi( \vect{\tau} | \vect{s}, o^{*} )$\\
			Record the data $\mathcal{D}_{\textrm{new}} = \{ (\vect{s}_{j}, \vect{\tau}_{j}, R_{j}) \}$\\
			Store the data $\mathcal{D} \leftarrow \mathcal{D} \cup \mathcal{D}_{\textrm{new}}$\\
		} 
		\UNTIL{ the task learned}
	\end{algorithmic}
	\label{alg:HPSDE}
\end{algorithm}

While the marginal distribution $p(o | \vect{s})$ obtained in the density estimation can be used as the gating policy in our framework, the gating policy presented in this section works better in practice as shown in the experimental result section.
In the experiment, we used the UCB policy to encourage the exploration in the selection of option policies~\cite{Auer03}. 

\subsection{Algorithm}
We summarize the procedure of HPSDE in Algorithm~\ref{alg:HPSDE}. 
An open parameter of HSPDE is the maximum number of option policies $O_{\textrm{max}}$. 
If sufficiently large $O_{\textrm{max}}$ is given, the number of the option policies is automatically determined.
If necessary, one can perform dimensionality reduction before the return-weighted density estimation step.




\subsection{Connection to Policy Parameter Estimation}
In prior work by \citeauthor{Daniel16}~(\citeyear{Daniel16}), the parameter of the option policy $\pi( \vect{\tau} | \vect{s}, o )$ needs to be estimated in order to estimate $p( o | \vect{s}, \vect{\tau})$.
On the contrary, our latent variable estimation does not need to estimate $\pi(\vect{\tau} | \vect{s}, o)$,
since our approach estimates the latent variable $o$ by directly approximating the density $p^{*}(\vect{s}, \vect{\tau})$.

On the other hand, our approach is closely related to prior work on estimating policy parameters for learning a monolithic policy.
Here, we consider a policy $\pi_{\vect{\theta}}$ parameterized by a vector $\vect{\theta}$.
In episodic policy search,
$D_{\textrm{KL}}( p^{*}(\vect{s}, \vect{\tau}) || p_{\vect{\theta}}(\vect{s}, \vect{\tau} ) )$ can be rewritten by using $p^{*}(\vect{s}, \vect{\tau}) = d(s)\pi^{*}(\vect{\tau} | \vect{s}) $ and $p_{\pi}(\vect{s}, \vect{\tau} ) = d(\vect{s})\pi_{\vect{\theta}}( \vect{\tau} | \vect{s} )$ as
\begin{align}
&D_{\textrm{KL}}( d(s)\pi^{*}(\vect{\tau} | \vect{s}) || d(s)\pi_{\vect{\theta}}(\vect{\tau} | \vect{s}) ) \nonumber \\
& = \iint d(\vect{s})\pi^{*}(\vect{\tau} | \vect{s})\log \frac{d(s)\pi^{*}(\vect{\tau} | \vect{s})}{ d(s)\pi_{\vect{\theta}}(\vect{\tau} | \vect{s})} \textrm{d}\vect{\tau}\textrm{d}\vect{s} \nonumber \\
& = \iint d(\vect{s})W(\vect{s}, \vect{\tau})\pi_{\vect{\theta}_{\textrm{old}}}(\vect{\tau} | \vect{s})  \log \frac{ W(\vect{s}, \vect{\tau})\pi_{\vect{\theta}_{\textrm{old}}}}{ \pi_{\vect{\theta}}(\vect{\tau} | \vect{s}) } \textrm{d}\vect{\tau}\textrm{d}\vect{s} \nonumber \\
&  \approx \frac{1}{N} \sum_{ (\vect{s}_{i}, \vect{\tau}_{i}) \in \mathcal{D} } \left( - W(\vect{s}_{i}, \vect{\tau}_{i})  \log \pi_{\vect{\theta}} (\vect{\tau} | \vect{s}) \right. \nonumber \\
& \qquad \left. + W(\vect{s}, \vect{\tau})\log W(\vect{s}, \vect{\tau})\pi_{\vect{\theta}_{\textrm{old}}} (\vect{\tau} | \vect{s}) \right).
\end{align}
Since the second term is independent of $\vect{\theta}$, the minimizer of $D_{\textrm{KL}}( d(s)\pi^{*}(\vect{\tau} | \vect{s}) || d(s)\pi_{\vect{\theta}}(\vect{\tau} | \vect{s}) )$ can be obtained by maximizing the weighted log likelihood as
\[
\pi^{*}_{\vect{\theta}} = \arg \max \frac{1}{N} \sum_{ (\vect{s}_{i}, \vect{\tau}_{i}) \in \mathcal{D} }  W(\vect{s}_{i}, \vect{\tau}_{i})  \log \pi_{\vect{\theta}} (\vect{\tau} | \vect{s}).
\]
This is exactly the episodic version of  reward-weighted regression~(eRWR)~\cite{Peters07,Kober11}.
Thus, the option estimation in our approach is closely related to eRWR for estimating the policy parameter, although our option estimation does not require estimating the option policy parameters.


\section{Experimental Results}
\label{sec:result}
To visualize the performance, we first evaluate HPSDE in toy problems and the puddle world task where the return functions are multi-modal.
Subsequently, we show the experiments with the motion planning task for a robotic manipulator, which is a practical application of hierarchical RL.
In the experiment, we evaluated variants of HPSDE:
we implemented HPSDE using REPS and reward weighted regression~(RWR)~\cite{Peters07} for updating the option policies.
REPS is a model-free policy search algorithm and constrains the KL divergence between the old and the updated policies in the policy update, although RWR does not have such a constraint in the policy update.
The constraint of the KL divergence on the policy update is frequently used to achieve stable exploration~\cite{Deisenroth13}.
We can see how the constraint on the option policy update influences on the performance of HPSDE by comparing HPSDE with REPS and HPSDE with RWR.
To evaluate the gating policy using a GP described in the previous section, we implemented a variant of HPSDE where the gating policy is represented by a softmax function, which is equivalent to the gating policy used by~\citeauthor{Daniel16}~(\citeyear{Daniel16}).
In addition, we compare the performance of HPSDE with HiREPS~\cite{Daniel16}, which is considered as one of the state-of-the-art model-free hierarchical policy search methods.
In the experiment, we assumed that a trajectory $\vect{\tau}$ can be represented by a trajectory parameter $\vect{\xi}$, and 
the task is to learn a policy $\pi( \vect{\xi} | \vect{s} )$ that generates trajectory parameter instead of a raw representation of a trajectory $\vect{\tau}$.
we used a Gaussian policy for an option policy given by
\begin{align}
\pi\left( \vect{\xi} | \vect{s}, o \right) = \mathcal{N}( \vect{\xi} | \vect{w}_{o}^{\top}\vect{\phi}(\vect{s}), \vect{\Sigma}_{o} ),
\end{align}
where $\vect{w}_{o}$ and $\vect{\Sigma}_{o}$ are option policy parameters to be learned, and the mean of the policy $\vect{w}_{o}^{\top}\vect{\phi}(\vect{s})$ is linear to the feature function $\vect{\phi}(\vect{s})$.
For computing the importance weight in \eqref{eq:normalized_w},
we used $f(R) = \exp(R)$.
We performed each task 20 times to evaluate the variance of the achieved return.
To deal with high-dimensional and non-Gaussian data, we employed the Laplacian eigenmaps~\cite{Belkin03} for nonlinear dimensionality reduction.

\subsection{Toy Problem}
\begin{figure}[t]
	\centering
	\subfigure[]{
		\includegraphics[width=0.305\columnwidth]{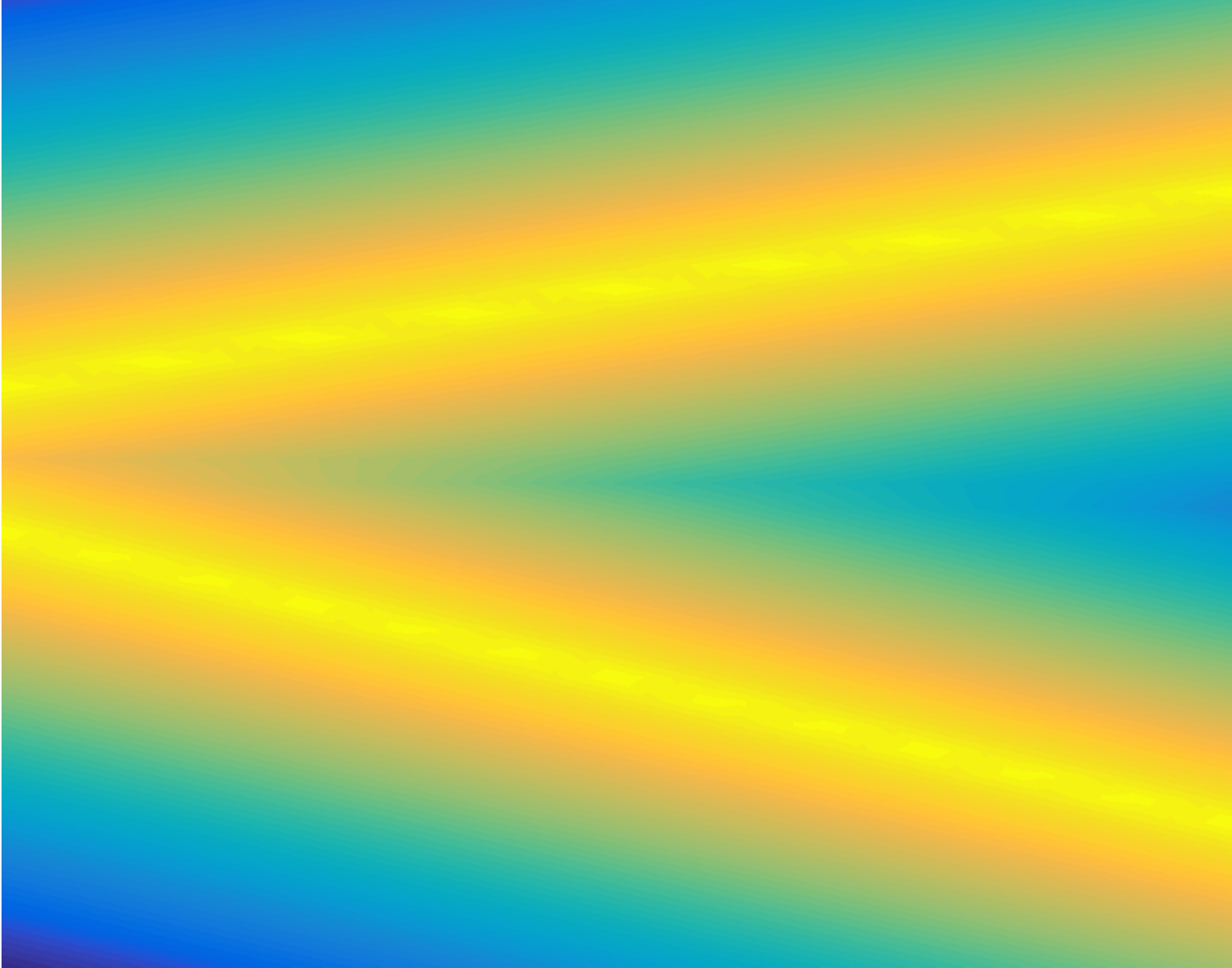}
	}
	\subfigure[]{
		\includegraphics[width=0.305\columnwidth]{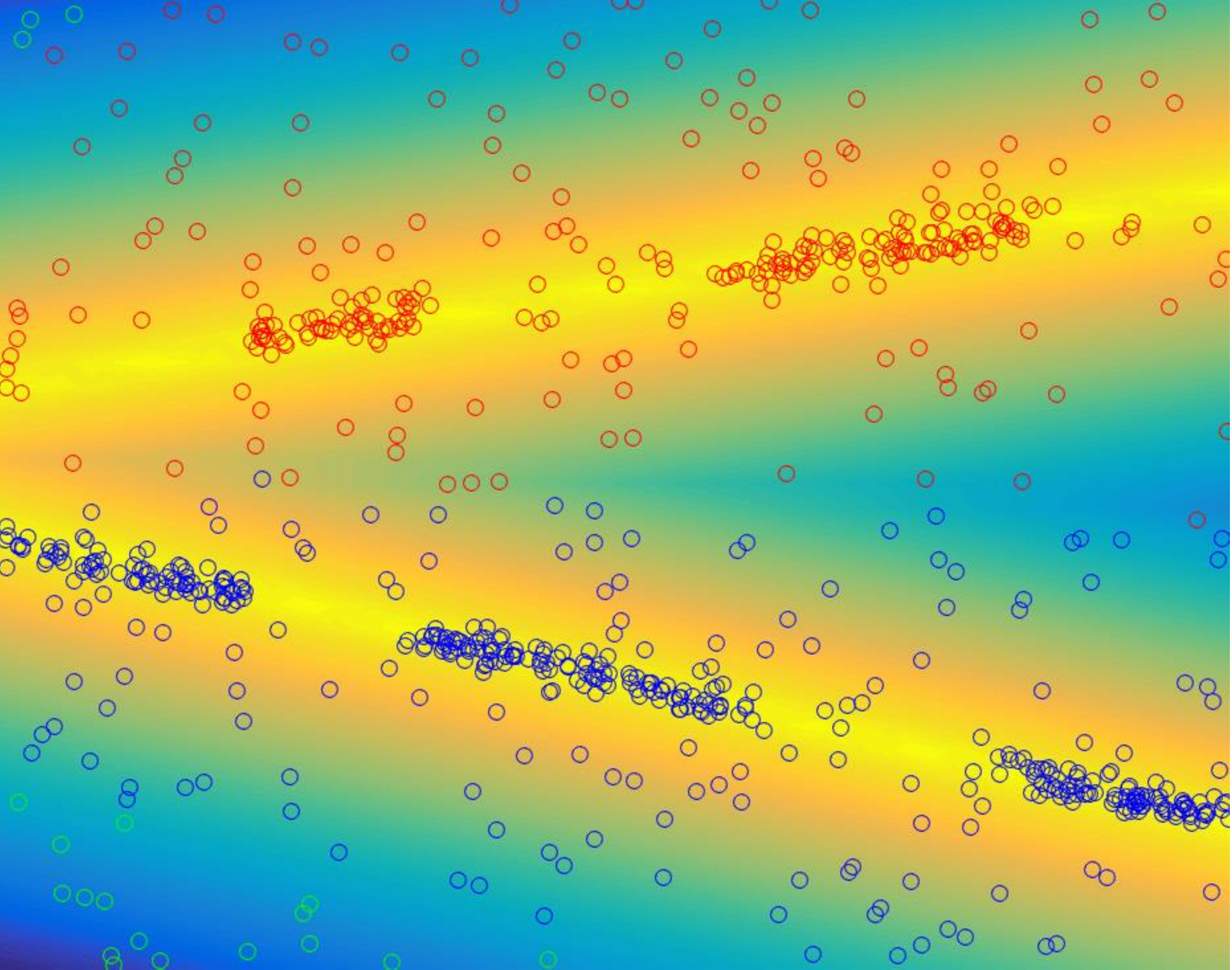}
	}
	\subfigure[]{
		\includegraphics[width=0.305\columnwidth]{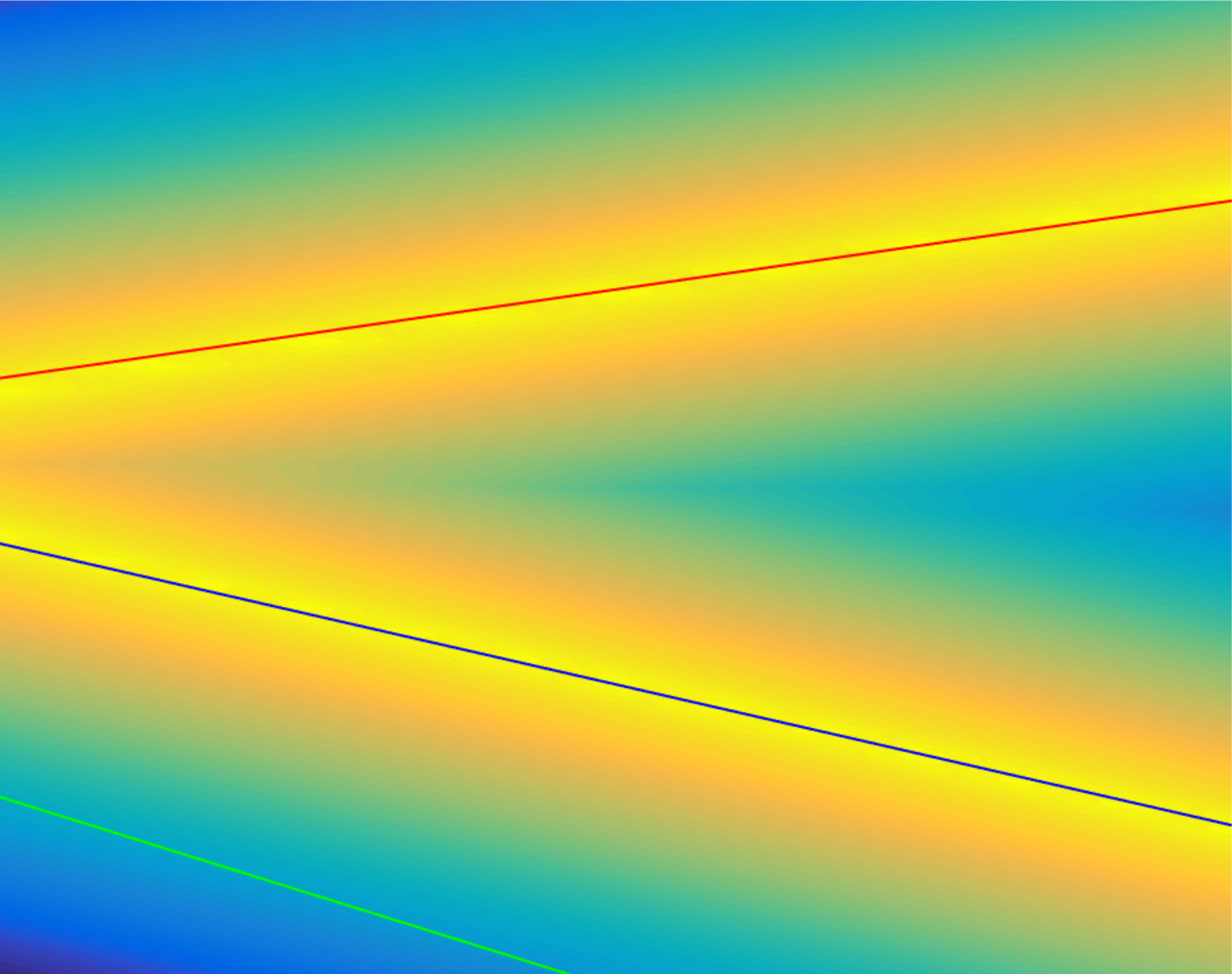}
	}
	\subfigure[]{
		\includegraphics[width=0.305\columnwidth]{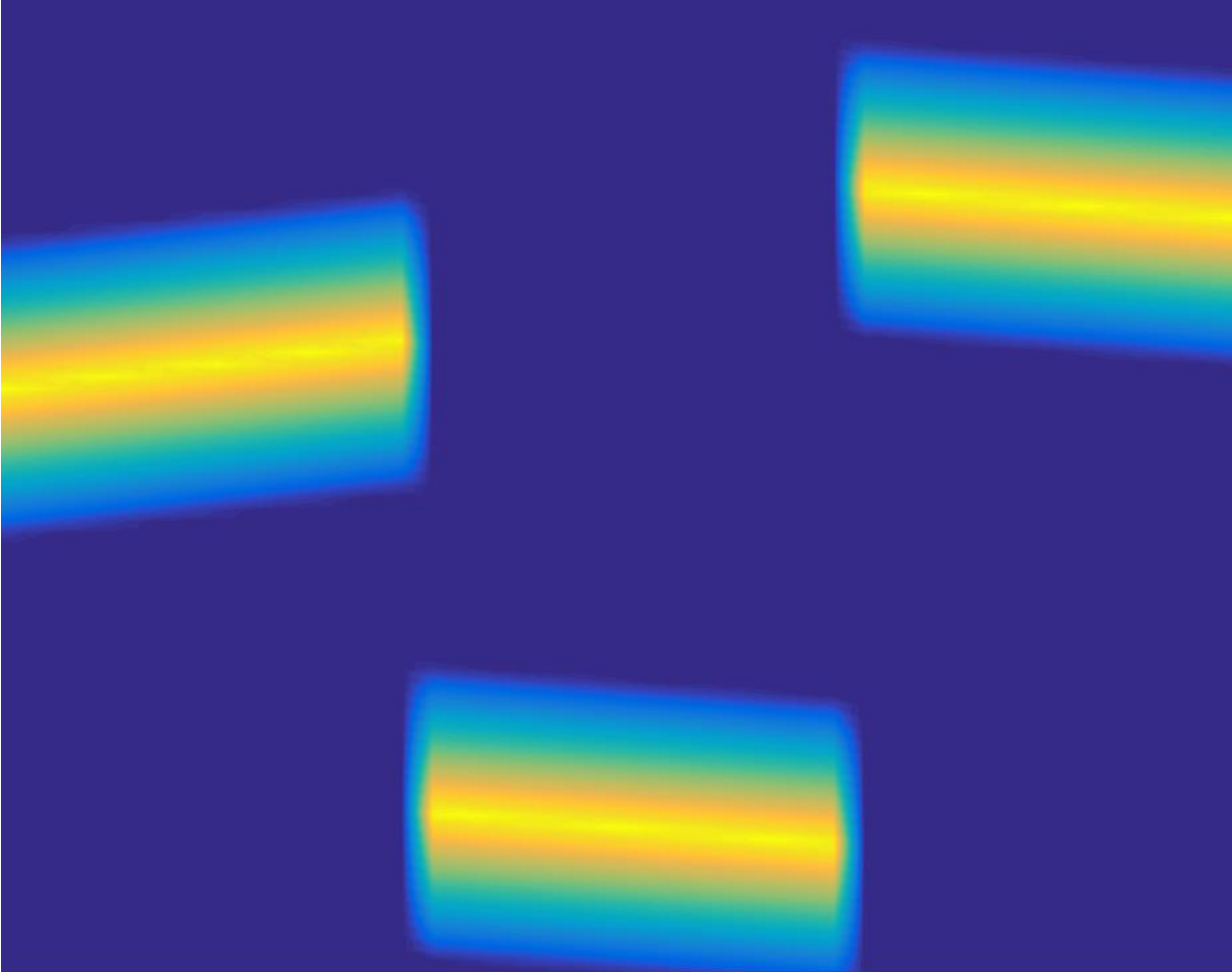}
	}
	\subfigure[]{
		\includegraphics[width=0.305\columnwidth]{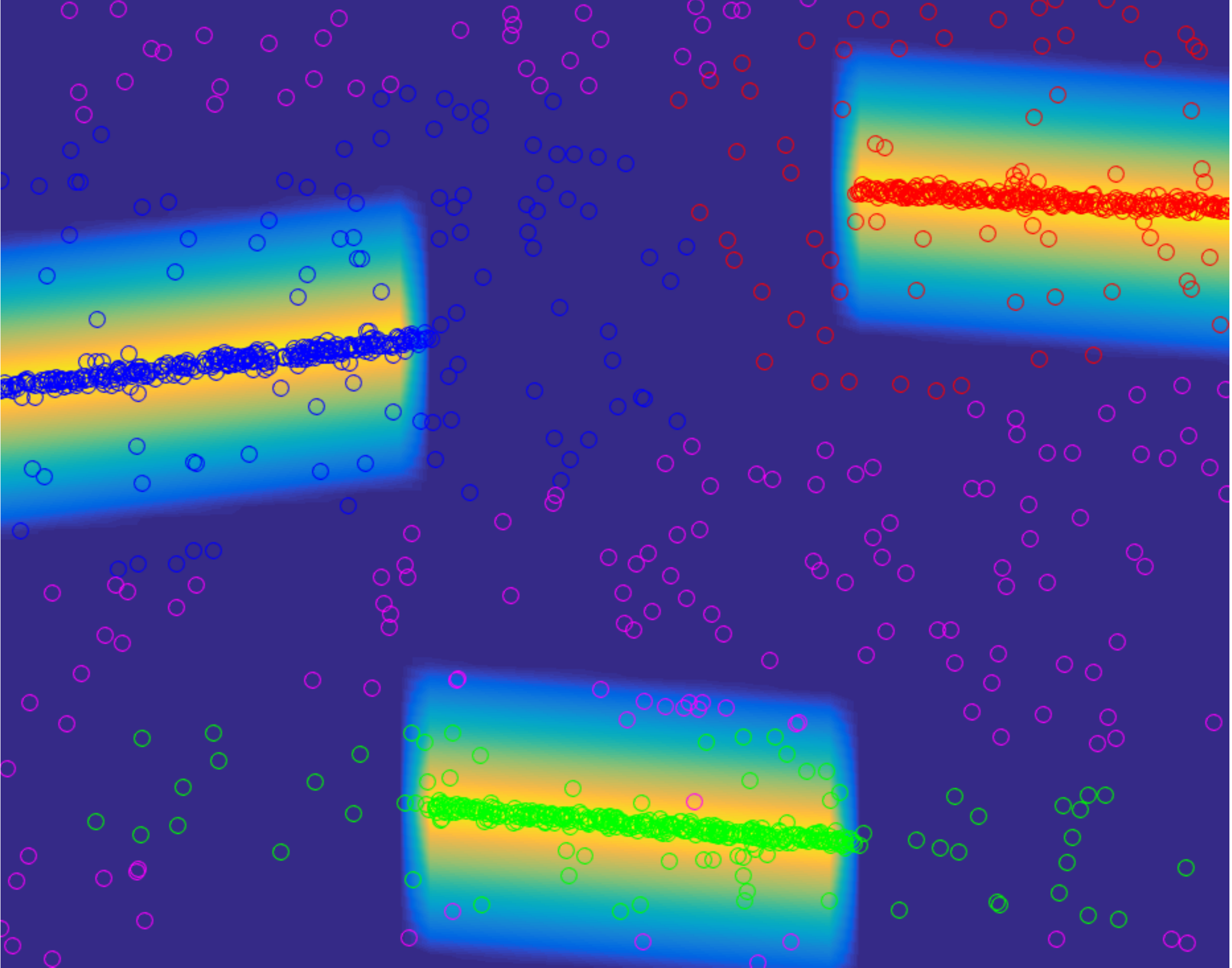}
	}
	\subfigure[]{
		\includegraphics[width=0.305\columnwidth]{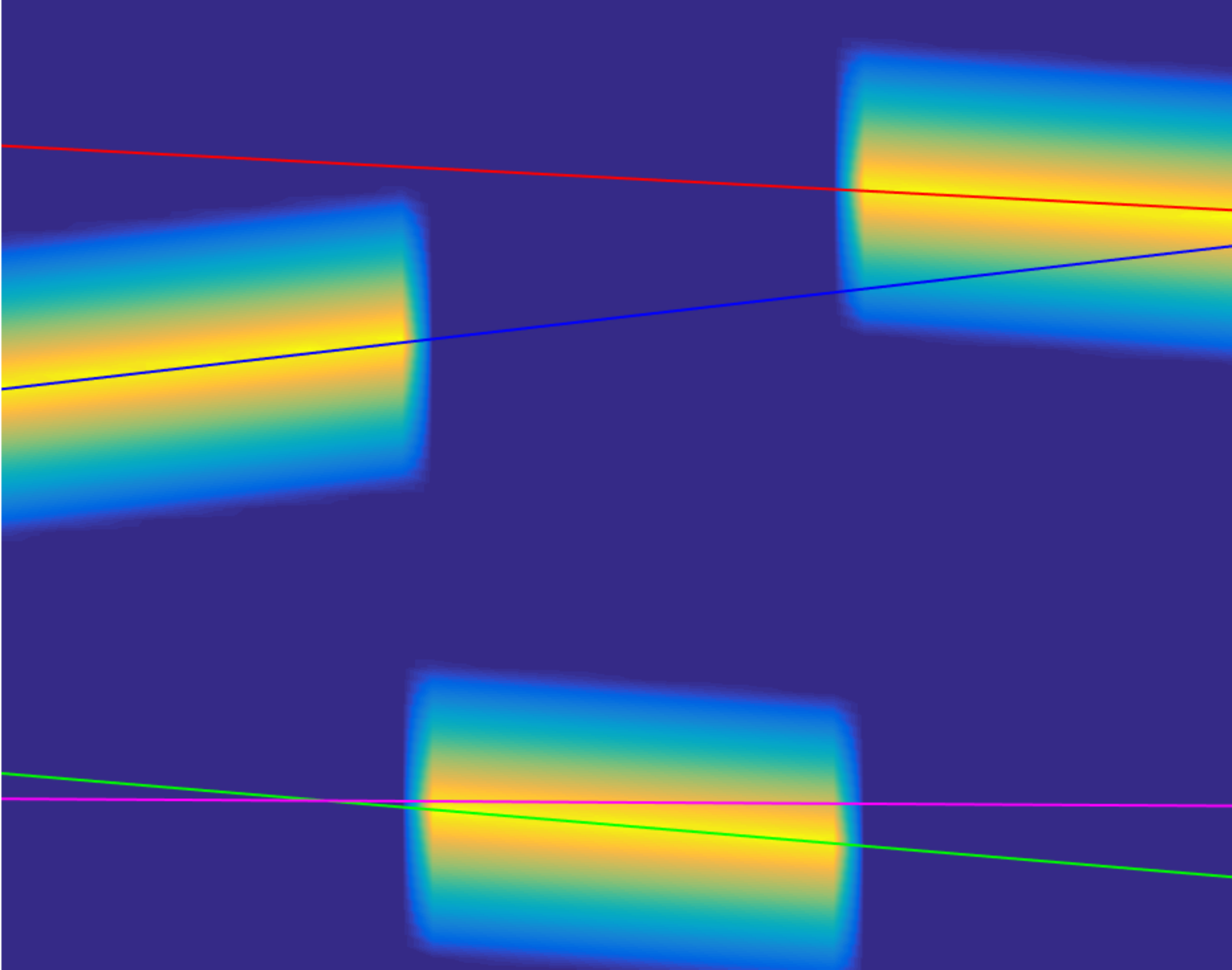}
	}
	\caption{ The toy problem using the return function with multiple modes. The horizontal axis shows the context, the vertical axis shows the trajectory parameter. The warmer color represents the higher return. (b) and (e) visualize samples collected by HPSDE.
		Learned option policies are visualized as lines as shown in (c) and (f). These results were obtained by HPSDE with REPS and the GP gating policy.
	} 
	\label{fig:toy}
\end{figure}

We first evaluate the performance of HPSDE in toy problems.
In this toy example, the context and the trajectory parameter are one-dimensional so that we can visualize the result intuitively. 
We performed evaluation using return functions with two and three modes, which are shown in Figure~\ref{fig:toy}.
For this task, we used a linear feature function $\vect{\phi}(\vect{s}) = [ \vect{s}^{\top}, 1 ]^{\top}$ and set $O_{\textrm{max}} = 10$ for HPSDE.

\begin{figure}[t]
	\centering
	\subfigure[]{
		\includegraphics[width=0.45\columnwidth]{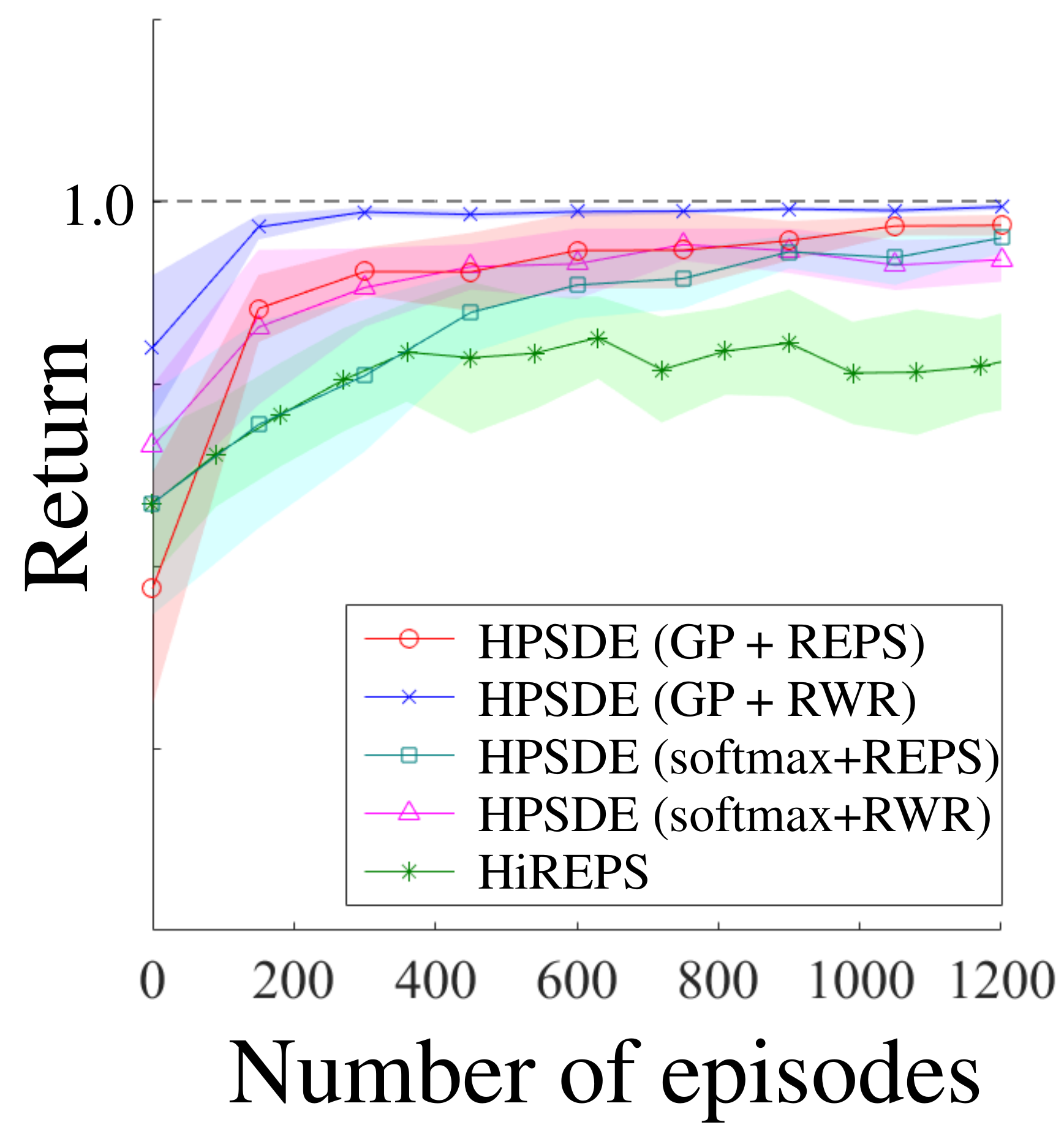}
	}
	\subfigure[]{
		\includegraphics[width=0.45\columnwidth]{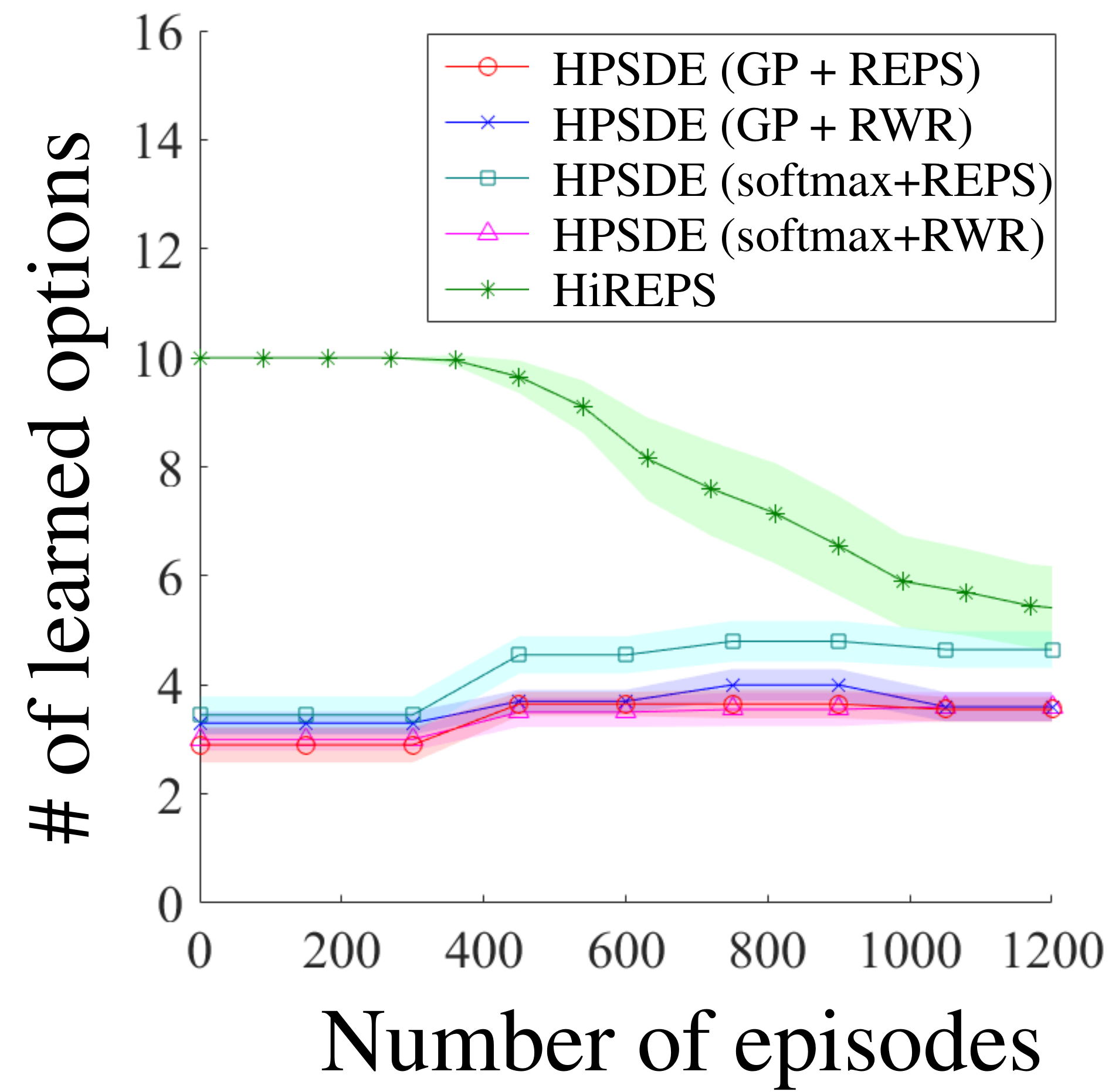}
	}
	\subfigure[]{
		\includegraphics[width=0.45\columnwidth]{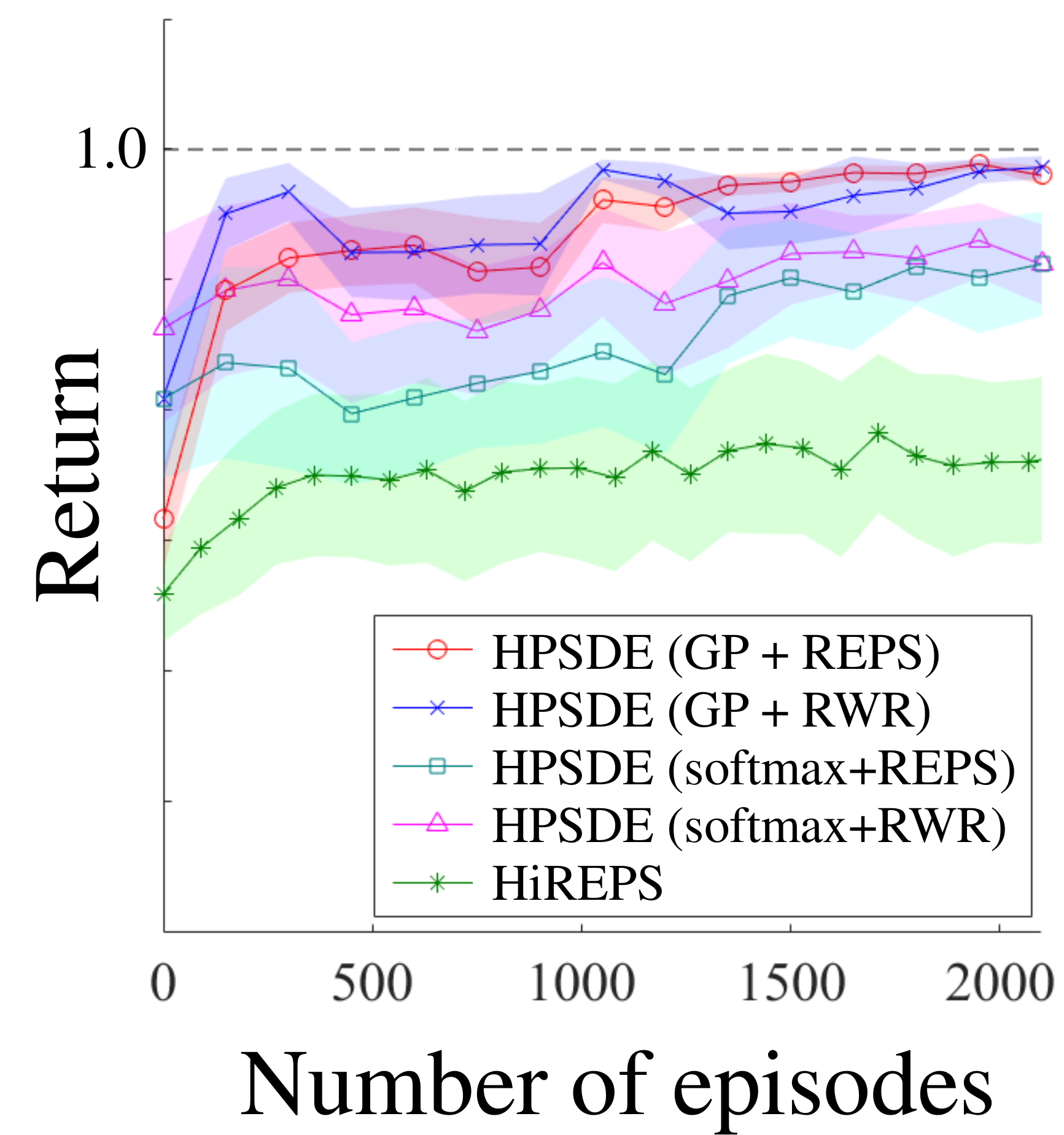}
	}
	\subfigure[]{
		\includegraphics[width=0.45\columnwidth]{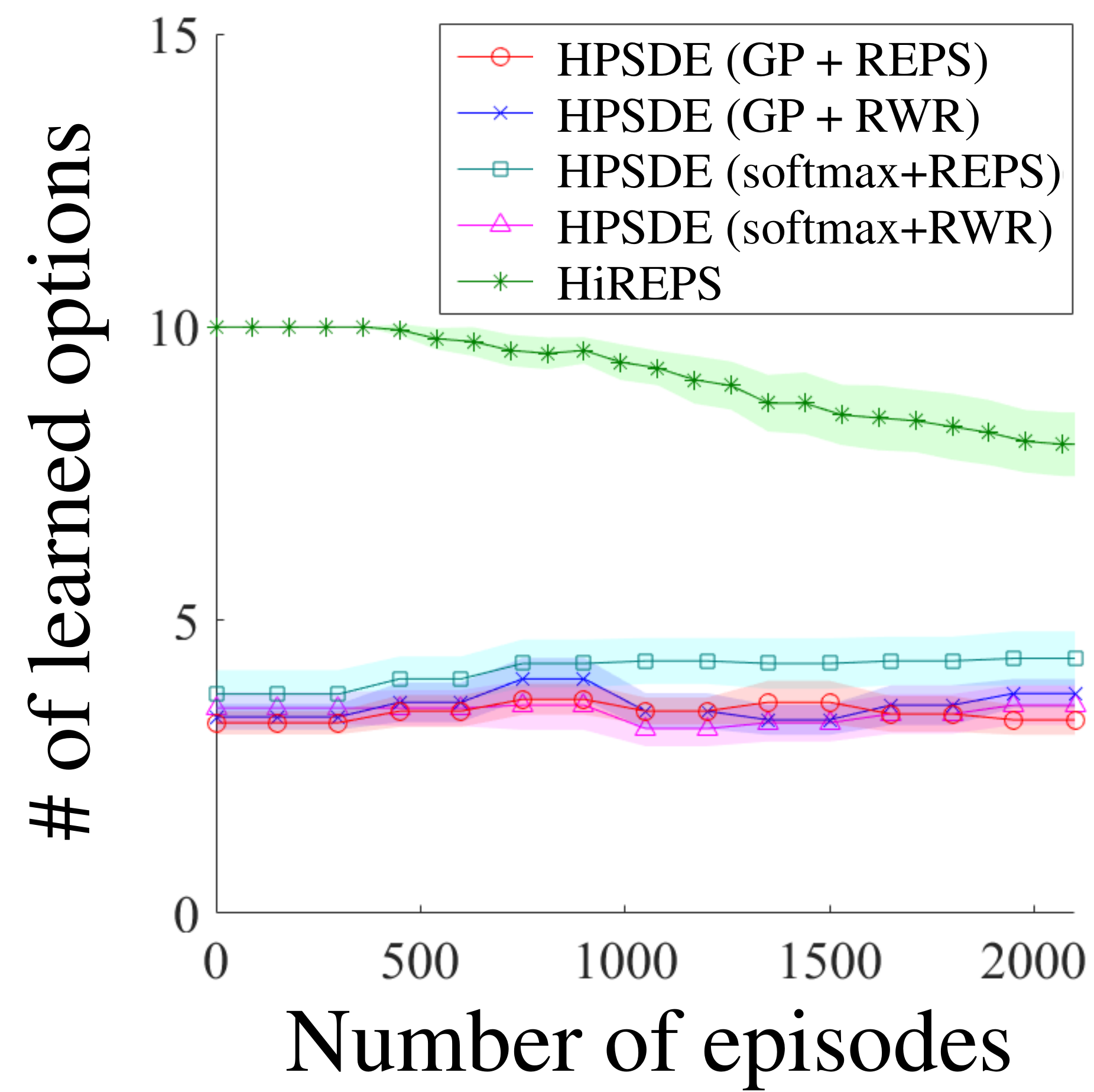}
	}
	\caption{ Results of the toy problems. (a) and (b) show the results with the return function that has two modes shown in Figure \ref{fig:toy}~(a)-(c),
		and (c) and (d) show the results with the return function that has three modes shown in Figure \ref{fig:toy}~(d)-(f).
	} 
	\label{fig:toy_result}
\end{figure}

Figure \ref{fig:toy} visualizes the samples collected in the learning process and policies learned by HPSDE.
As shown, HPSDE identified the modes of the return function in both toy problems.
Although our approach extracted option policies from sample clusters with low returns, 
it is rarely selected since the expected return is always lower than other  policies.

The resulting return and the number of the learned policies are shown in Figure~\ref{fig:toy_result}. HPSDE achieved higher returns than HiREPS in these toy tasks. 
In addition, although both HiREPS and HPSDE converge to comparable numbers of option policies, HPSDE optimizes the number of the option policies much faster.
Since the policy updates of option policies in HiREPS and HPSDE with REPS are equivalent, 
this result indicates that the identification of the option policy structure in HPSDE is more efficient than that in HiREPS.
In addition, although HiREPS cannot increase the number of the option policies in the learning process,
our approach finds the optimal number of the option policies without such a limitation.

With regard to the gating policy, the gating policy with the return approximation using a GP outperforms the gating policy represented by a softmax function.
With respect to the strategy for updating the option policies, the learning rate of HPSDE with RWR was comparable to that of HPSDE with REPS in the toy problems, although REPS was reported to achieve faster learning than RWR in many cases~\cite{Peters10}. 
When a given sample distribution has multiple modes of the return function, 
the constraint on the KL divergence between the old and the updated policies prevents from jumping from the current mode to another mode.
On the other hand, when a sample distribution given to a policy has just one mode of the return function, 
the constraint in the policy update does not clearly improve the learning performance.
Since our framework successfully identifies the modes of the return function in these toy examples, 
the constraint on the KL divergence in the policy update does not make a clear difference of performance between REPS and RWR in HPSDE.

\begin{figure}
	\centering
	\includegraphics[height=0.15\textheight, width=\columnwidth]{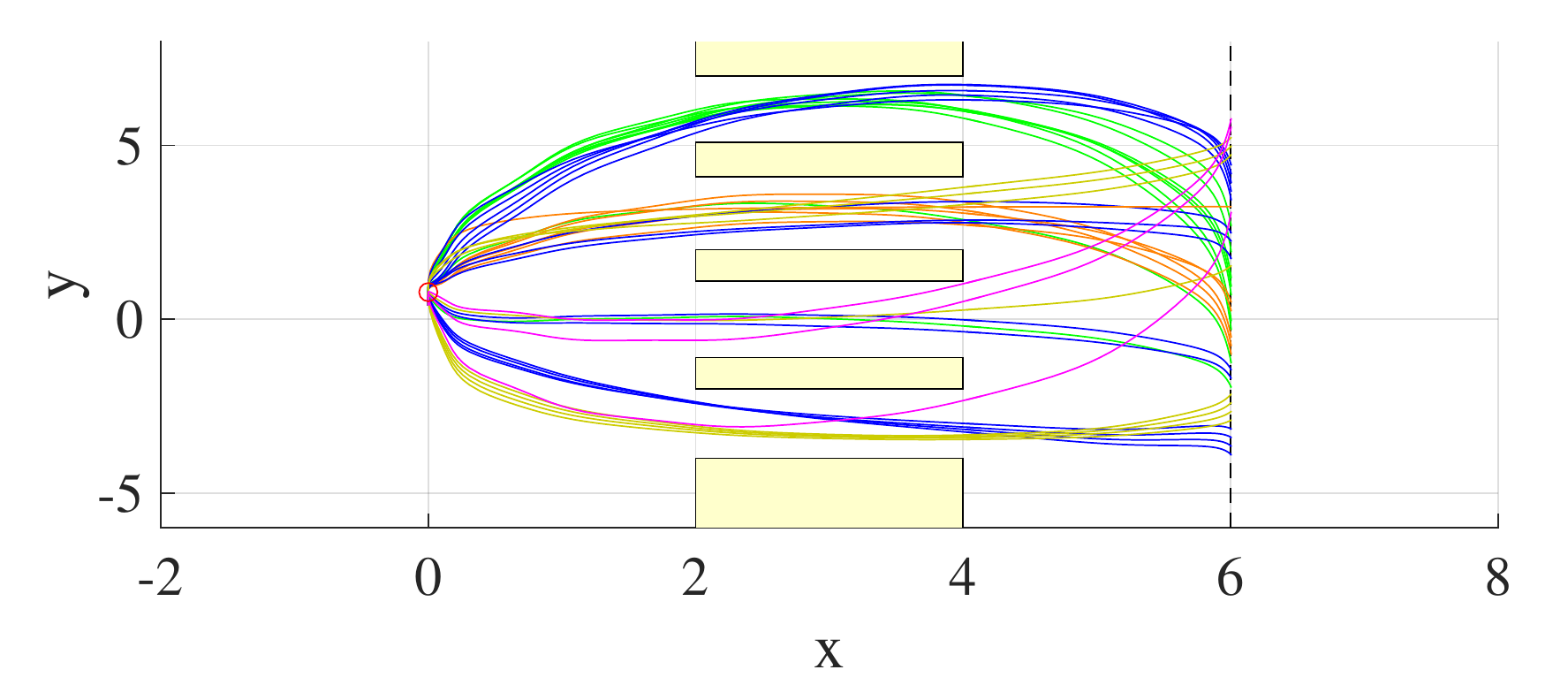}
	\caption{The puddle world task. Routes from the starting position to the goal positions are found by HPSDE. 
		The starting position is fixed as shown as the red point and the goal positions are distributed on the line $x = 6$. Trajectories in the same color are generated by the same option policy. 
	}
	\label{fig:dmp}
\end{figure}

\subsection{Puddle World Task}
We tested HPSDE with a variant of the puddle world task reported by \citeauthor{Daniel16}~(\citeyear{Daniel16}) where a collision free trajectory needs to be planned in a continuous space.
We used DMPs to represent trajectories in this task~\cite{Ijspeert02}.
The arrangement of the puddles is shown in Figure~\ref{fig:dmp}, which is more complex than the task reported by \citeauthor{Daniel16}~(\citeyear{Daniel16}).
We assume that the starting position is fixed, and the goal position, which is given as a context of this task, is distributed on the line $x = 6$. 
A trajectory is represented with two DMPs, which represent x and y dimensions, respectively.
In this task, the DMP for x dimension is fixed and the DMP for y dimension is optimized with HPSDE. 
We use 10 basis functions to represent a trajectory, and therefore HPSDE optimized 10 parameters of the DMP.
The return function is designed such that passing through the puddle results in a negative return and a longer trajectory receives a less return, which encourages a shorter and collision-free trajectory. 
For this experiment, we used the squared exponential feature: 
\begin{align}
\phi_{i}(\vect{s}) = \exp\left( (\vect{s} - \vect{s}_{i})^{\top} \Lambda (\vect{s} - \vect{s}_{i})  \right) ,
\end{align}
where $\Lambda$ is a diagonal matrix that defines the bandwidth.
This exponential feature enables us to represent a nonlinear policy~\cite{Daniel16}.
We set $O_{\textrm{max}} = 20$ for HPSDE.

\begin{figure}[]
	\centering
	\subfigure[]{
		\includegraphics[width=0.45\columnwidth]{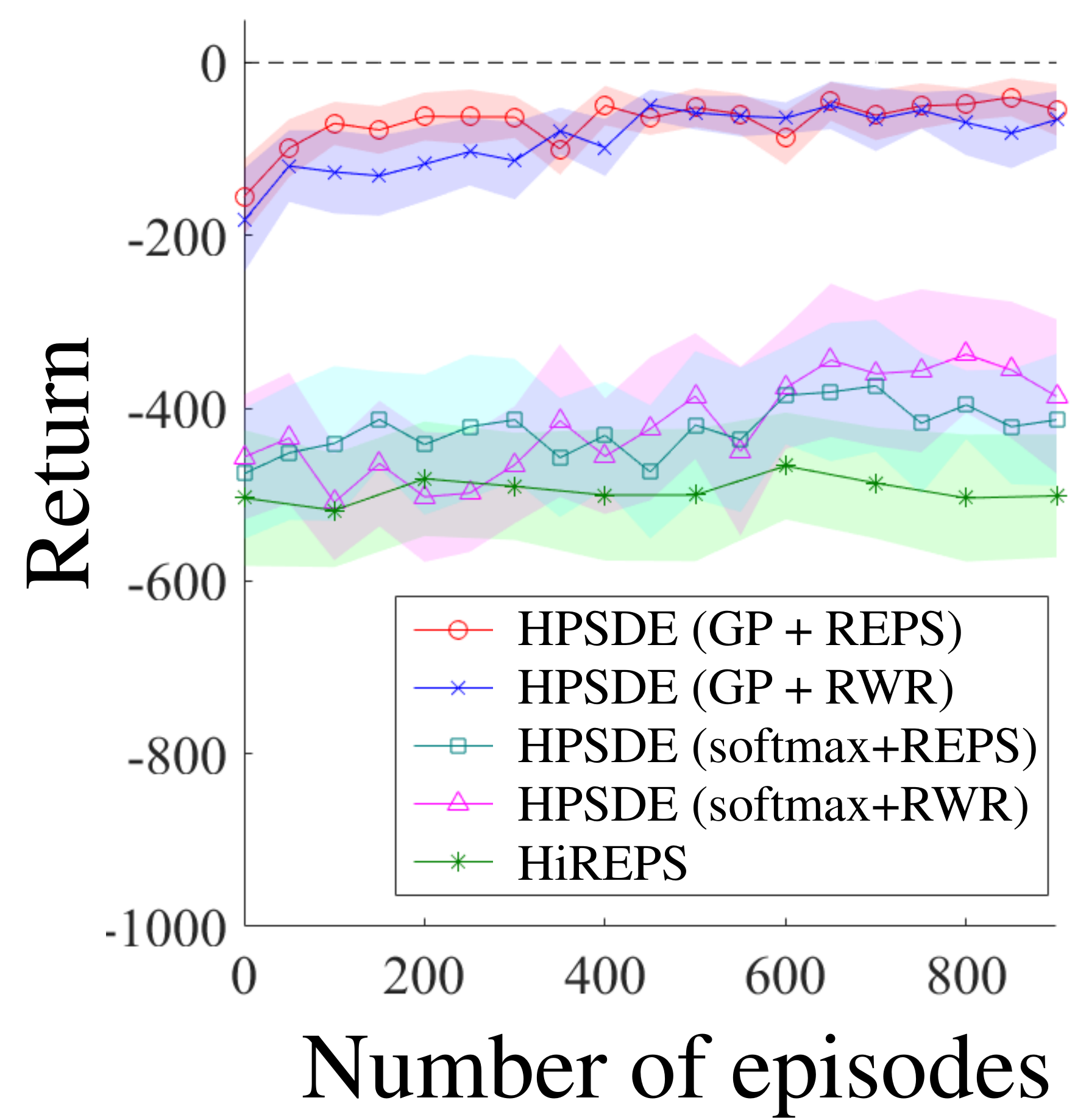}
	}
	\subfigure[]{
		\includegraphics[width=0.45\columnwidth]{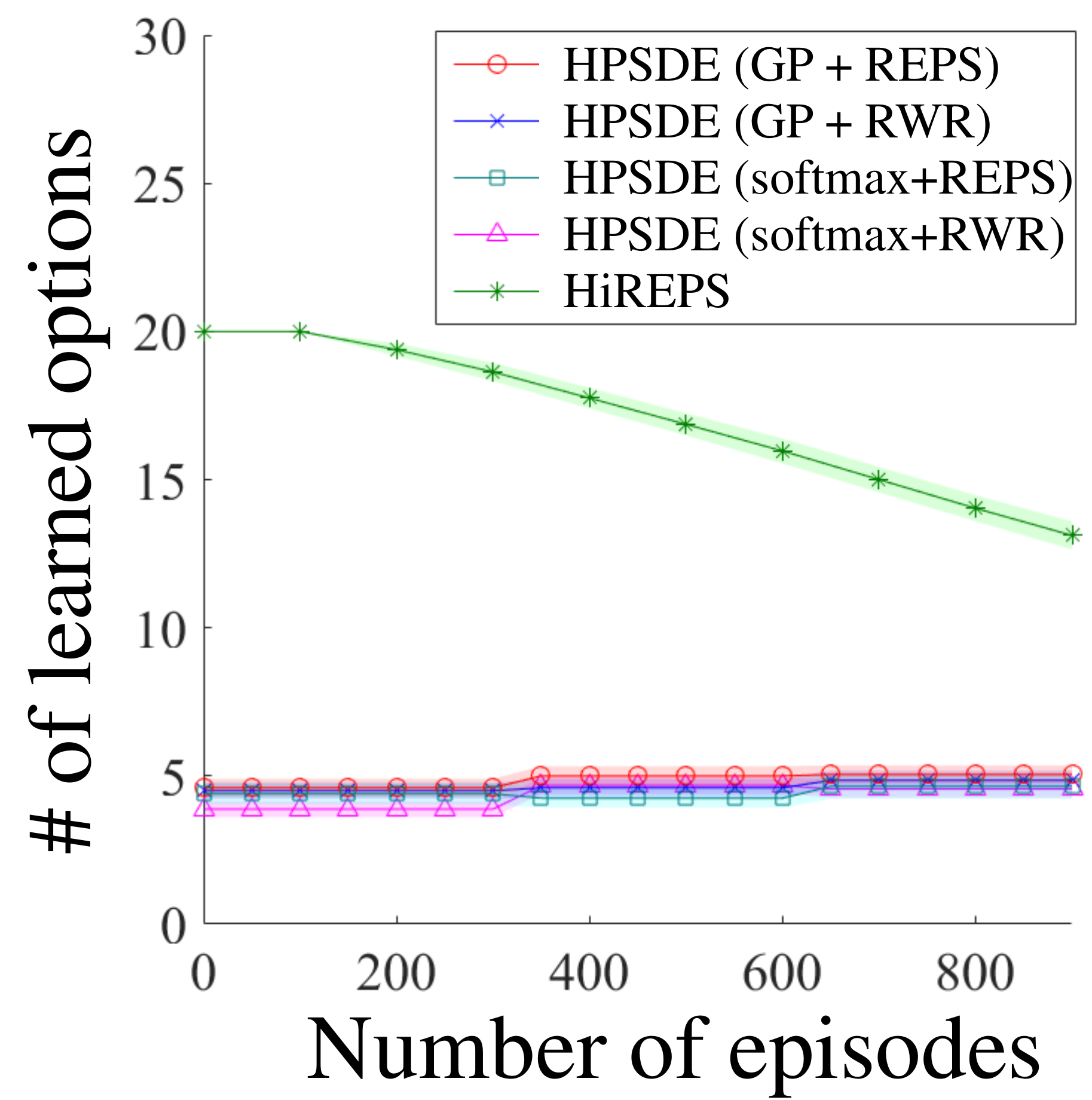}
	}
	\caption{Results of the puddle world task. (a) returns and (b) the number of the learned option policies.}
	\label{fig:dmp_result}
\end{figure}

As shown in Figure \ref{fig:dmp}, HPSDE finds multiple solutions in the puddle world task.
Figure \ref{fig:dmp_result} shows the resulting return and the number of the learned option policies.
In this experiment, HiREPS falls into local optima and converges to poor performance.
On the contrary, HPSDE achieves much higher performance by finding option policies corresponding to modes of the return functions. 
These results show that the learning of the option policy structure with HPSDE is more efficient than HiREPS.


\subsection{ Motion Planning for a Redundant Manipulator }
\begin{figure}
	\centering
	\includegraphics[width=\columnwidth]{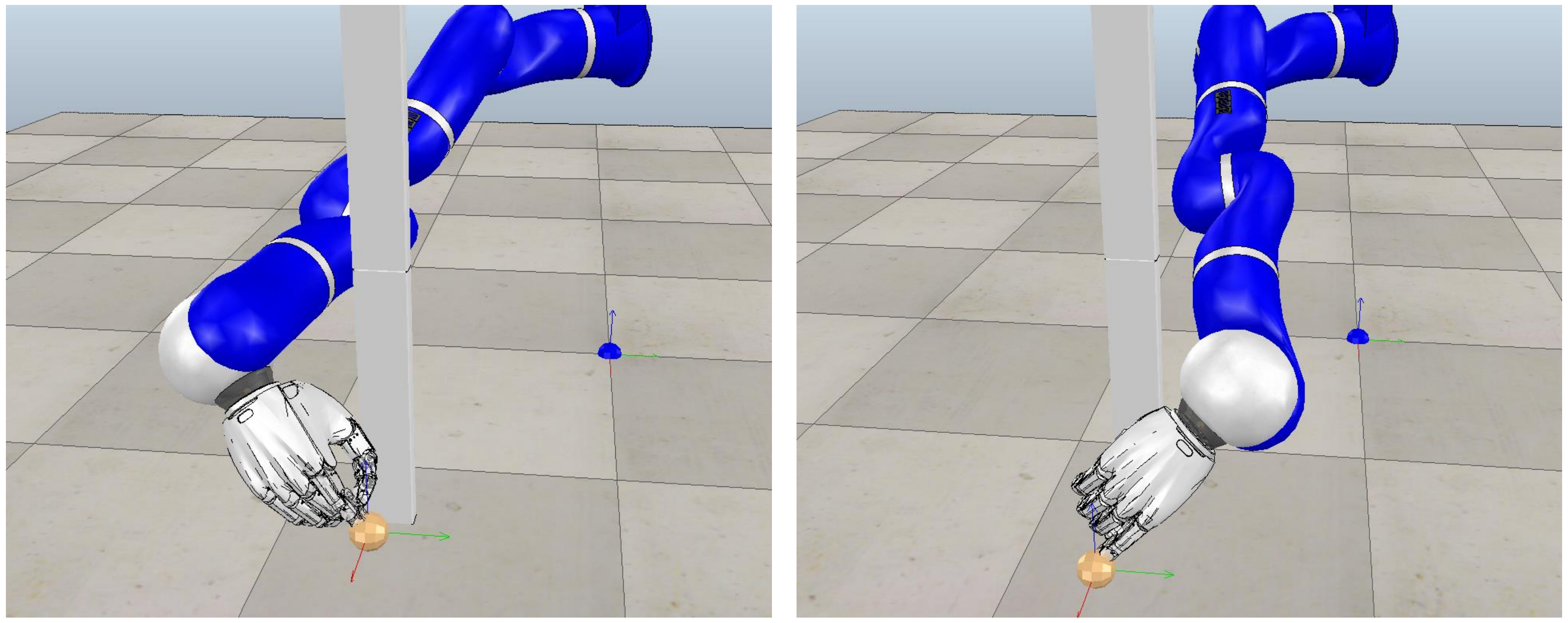}\\
	\caption{The reaching task with a robot with 7 degrees of freedom. Multiple postures to reach the desired point were learned by HPSDE.
		The orange sphere shows the goal point.
	}
	\label{fig:reaching_robot}
\end{figure}

Planning a motion of a redundant manipulator has been an open problem in robotics since 
there exists multiple solutions in the continuous space. 
We evaluated HPSDE with a motion planning problem for such a redundant manipulator in a simulation environment, developed based on V-REP~\cite{Rohmer13}.
Figure~\ref{fig:reaching_robot} shows the simulation environment.
A KUKA Light Weight Robot with 7 degrees of freedom is modeled in this simulation, and the task is to touch a desired point behind a pole. 
The goal point is distributed behind the pole, and the context of this task is given by the Cartesian coordinates of the goal position as $\vect{s} = [ x, y, z ]$. 
In this task, the objective of hierarchical RL is to optimize the final configuration of the robotic manipulator $\vect{q} \in \Real^{7}$.
The return is given by $R(\vect{q}) = - d(\vect{q}) - C(\vect{q})$, where $d(\vect{q})$ is the distance between the position of the end-effector and the desired position, and $C(\vect{q})$ is the cost of colliding with the pole for a given configuration $\vect{q}$.
The collision cost is computed based on the cost function proposed by~\citeauthor{Zucker13}~(\citeyear{Zucker13}).
We used the squared exponential feature as in the previous experiment.
We set $O_{\textrm{max}} = 10$ for HPSDE.

\begin{figure}[t]
	\centering
	\subfigure[]{
		\includegraphics[width=0.465\columnwidth]{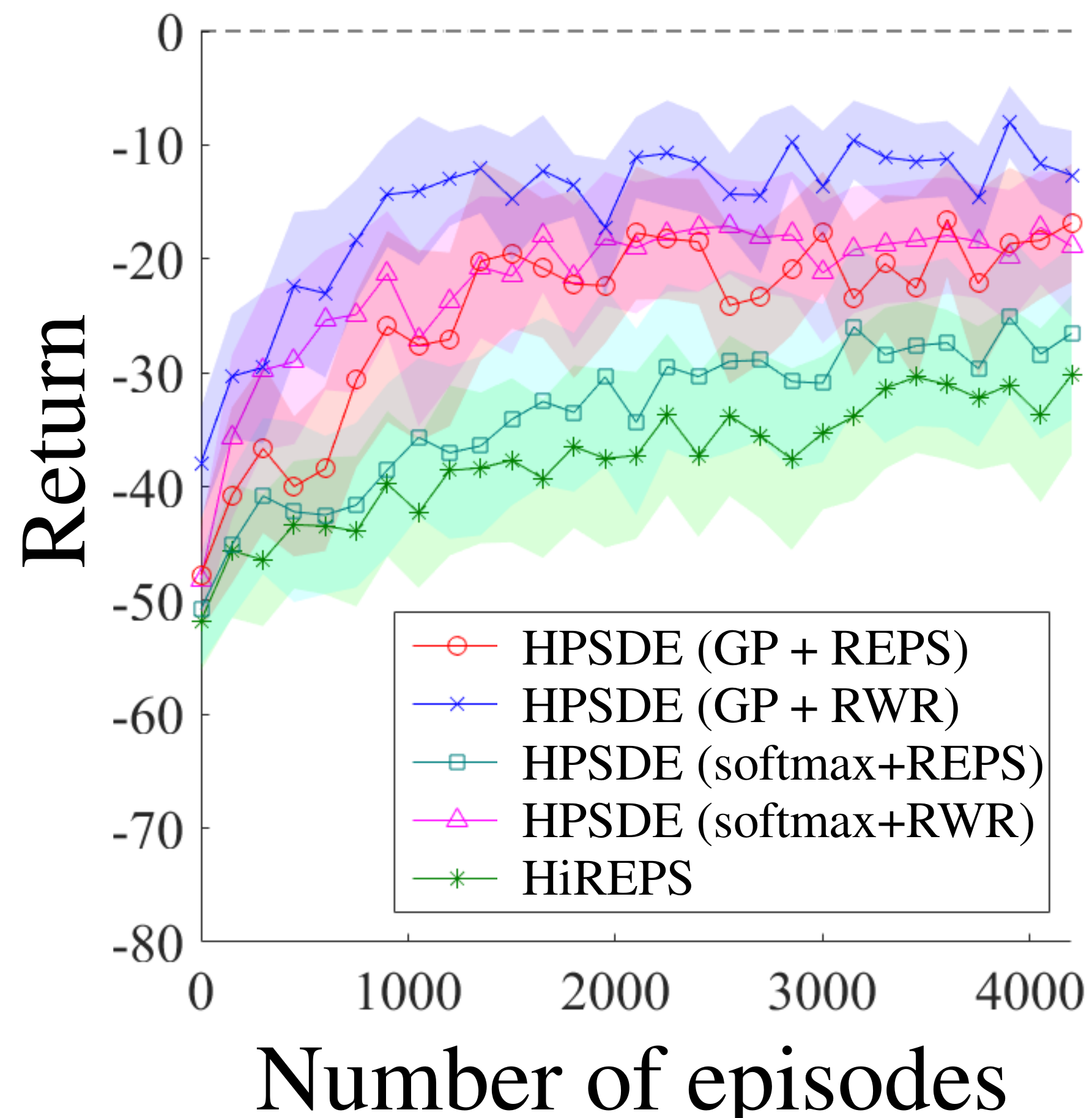}
	}
	\subfigure[]{
		\includegraphics[width=0.465\columnwidth]{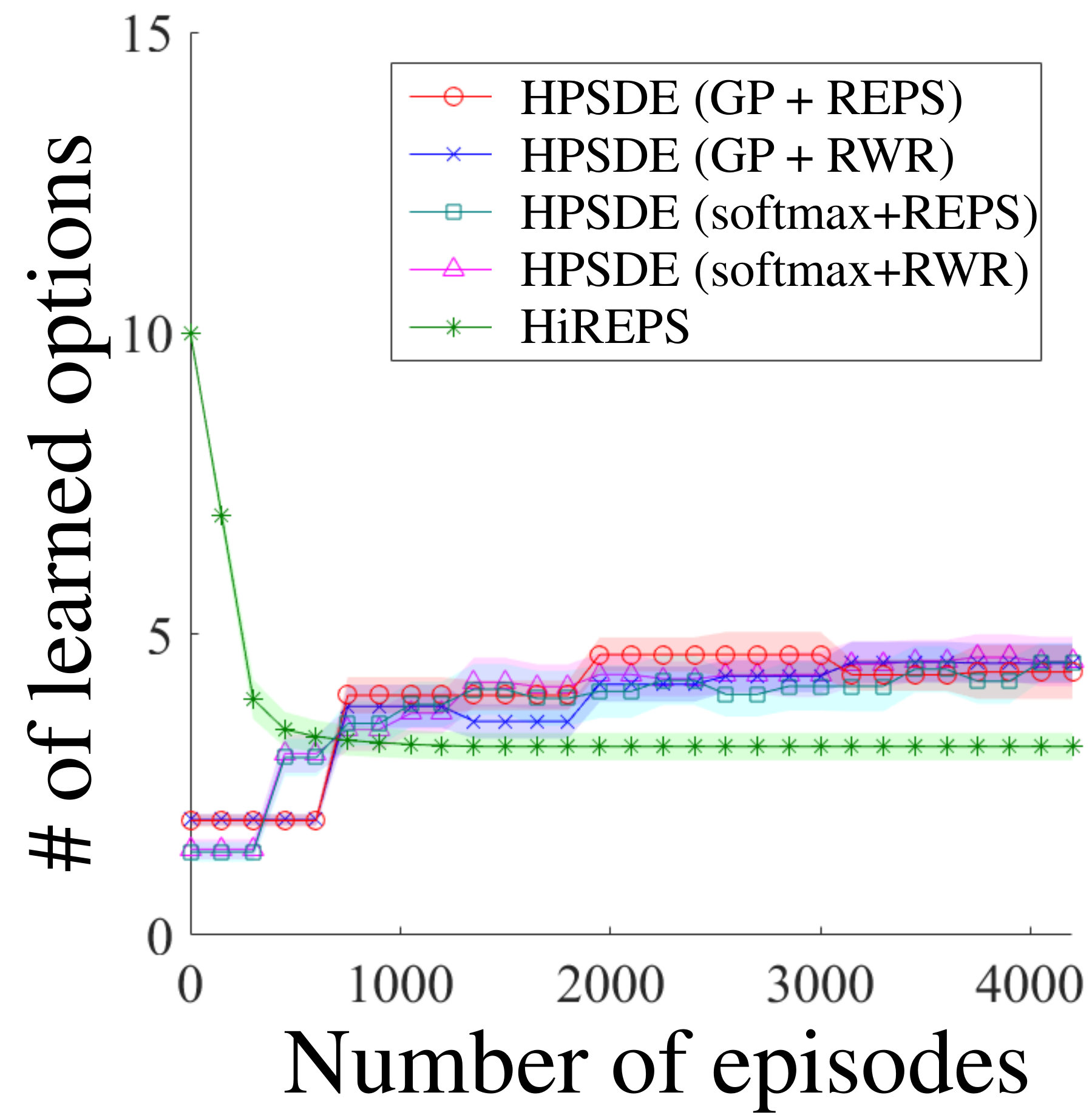}
	}
	\caption{Results of the motion planning task for a redundant manipulator. (a) returns and (b) the number of the learned option policies.}
	\label{fig:reaching_result}
\end{figure}

As shown in Figure~\ref{fig:reaching_robot}, HPSDE found multiple policies to achieve the reaching task.
The learning performance is shown in Figure~\ref{fig:reaching_result}.
HPSDE with REPS and the GP gating policy demonstrates the best performance in this task.  
Although HiREPS also achieves performance comparable to some variants of HPSDE, 
it is necessary to heuristically specify the minimum and maximum numbers of the option policies and the parameter to delete the option policies in order to obtain the best performance of HiREPS for this task.
On the contrary, an open parameter in HPSDE is just the maximum number of the option policies.

When the start and goal configurations are given, trajectory optimization methods developed in robotics such as CHOMP~\cite{Zucker13} and TrajOpt~\cite{Schulman14} can be used for motion planning.  
However, planning the desired configuration itself is often challenging since there exist multiple solutions in multi-dimensional and continuous space.
The result in this work indicates that HPSDE can address such motion planning problems in robotics well.

\section{Discussion}

Regarding the computation cost, our method is comparable to other methods 
in terms of executing an off-the-shelf policy search method for updating option policies.
On the other hand, since our method finds the optimal number of option policies, 
we can avoid using an unnecessarily large number of option policies, which would result in additional computation costs.
The computation cost for estimating the latent variable is much less than the cost for updating the policy parameters, 
since executing the EM algorithm is fast.
The most time-consuming part in our method is training the gating policy with a GP. 
Due to the limitation of a GP, it is hard to scale it to high-dimensional data.
The development of the gating policy that can deal with high-dimensional data such as image inputs will be our future work, which will be potentially addressed by using a deep learning approach.

\section{Conclusion}
\label{sec:conclusion}
We proposed the hierarchical policy search via return-weighted density estimation~(HPSDE).
To address the issue of determining the structure of the option policies, our approach reduces the problem of estimating the modes of the return function to the problem of estimating the return-weighted sample density with a mixture model.
HPSDE automatically identifies the option policy structure, where each option policy corresponds to a single mode of the return function.  
The connection between the expected return maximization and the return-weighted density estimation is analytically shown in this study.
The experimental results show that HPSDE outperforms a state-of-the-art method for hierarchical reinforcement learning,
and that HPSDE can be used to solve motion planning problem for a redundant robotic manipulator.
In future work, we will extend the proposed approach such that high-dimensional data such as image inputs can be incorporated.
Additionally, we will extend the episodic HPSDE proposed in this paper to the learning of an action-state level policy and perform experiments with a real robot.

\section{ Acknowledgments}
OT and MS were supported by KAKENHI 17H00757.

\bibliographystyle{aaai}
\bibliography{AAAI18}

\section{Appendix}
\subsection{Implementation Details of VBEM with Importance Weights}

Here we describe some details of  the VBEM algorithm with importance weights.
The basic implementation follows the algorithm described in \cite{Bishop06,Sugiyama15}.
For details of the VBEM algorithm, please refer to \cite{Bishop06,Murphy12}.

For a given samples $\mathcal{D} = \{\vect{x}_{i}\}_{i=1}^{n}$, we consider the mixture of $m$ Gaussian models:
\begin{align}
q( \vect{x} | \mathcal{K}, \mathcal{M}, \mathcal{S} ) = \sum_{\ell=1}^{m} k_{\ell} \mathcal{N}( \vect{x} | \vect{\mu}, \vect{S}^{-1} ),
\end{align}
where $\mathcal{K} = \{k_1,...,k_m \}$ is a set of mixture coefficients, $\mathcal{M}=\{ \vect{\mu}_1,...,\vect{\mu}_m \}$ is a set of the means, and $\mathcal{S}=\{\vect{S}_1,...,\vect{S}_m \}$ is a set of the precision matrices.
Subsequently, we consider latent variables $O = \{ o_1,...o_n \}$ and the variational distribution $q(O)q(\mathcal{K}, \mathcal{M}, \mathcal{S})$. 
We choose a Dirichlet distribution as a prior for the mixing coefficients $\mathcal{K}$ and a Gaussian-Wishart distribution as a prior for each Gaussian component. 
The goal of VBEM is to maximize the lower bound of the marginal log likelihood $\log p(\mathcal{D})$ by repeating processes called VB-E and VB-M steps.

In the VB-E step, we compute the distribution of the latent variable $O$ from the current solution as
\begin{align}
 q(O) = \prod_{i=1}^{n} \prod_{\ell = 1}^{m} \hat{\eta}^{o_{i \ell}}_{i,j}, 
\label{eq:VB-E}
\end{align}
where $\hat{\eta}_{i,\ell}$ is the responsibility of the sample $i$ on the $\ell$th cluster.  
The responsibility  $\hat{\eta}_{i,\ell}$ is given by 
\begin{align}
\hat{\eta}_{i,\ell}=\frac{ \hat{\rho}_{i,\ell'}}{\sum_{\ell'=1}^{m} \hat{\rho}_{i,\ell'}},
\end{align}
where $\hat{\rho}_{i,\ell'}$ is computed as
\begingroup\makeatletter\def\f@size{9}\check@mathfonts
\def\maketag@@@#1{\hbox{\m@th\large\normalfont#1}}
\begin{align}
& \hat{\rho}_{i,\ell'}  = \exp \left( \psi(\hat{\alpha}_{\ell}) - \psi\left( \sum_{\ell'=1}^{m} \hat{\alpha}_{\ell'} \right) +
\frac{1}{2}\sum_{j=1}^{d} \psi \left( \frac{\hat{\nu_{\ell}} + 1 - j}{2}  \right) \right. \nonumber \\
& + \frac{1}{2} \log \det( \hat{\vect{K}}_{\ell} ) - \frac{d}{2\hat{\beta}_{\ell}} - \frac{\hat{\nu_{\ell}}}{2} 
( \vect{x}_{i} - \hat{\vect{h}}_{\ell} )\hat{\vect{K}}_{\ell}( \vect{x}_{i} - \hat{\vect{h}}_{\ell} )^{\top}
\Bigg),
\label{eq:responsibility}	
\end{align}
\endgroup

\noindent and $\psi(\alpha)$ is the digamma function defined as the log-derivative of the gamma function.

\begin{algorithm}[t]
	\caption{VBEM with Importance Weights }
	\begin{algorithmic}
		\STATE{
			\textbf{Input:} Data samples $\mathcal{D} = \{\vect{x}_{i}\}_{i=1}^{n}$, importance weights $\vect{w} = \{w_{1},...,w_n\}$, and hyperparameters
			$\alpha_{0}$, $\beta_{0}$, $\nu_0$, and $\vect{K}_{0}$  \\
			Initialize parameters \\
		}
		\REPEAT
		\STATE{ 
			VB-E step: \\
			\ \ \ \  Compute the distribution of the latent variable $q(O)$ \\
			\ \ \ \  by updating the responsibility $ \{\hat{\eta}_{i, \ell}\}_{i=1,\ell=1}^{n, \ \ \ m}$ with\\
		    \ \ \ \  $ \{ \hat{\alpha}_{\ell}, \hat{\beta}_{\ell}, \hat{\vect{h}}_{\ell}, \hat{\nu}_{\ell}, \hat{\vect{K}}_{\ell} \}$ \\
			VB-M step: \\
			\ \ \ \  Compute the joint distribution $q(\mathcal{K}, \mathcal{M}, \mathcal{S})$ by \\ 
			\ \ \ \  updating $ \{ \hat{\alpha}_{\ell}, \hat{\beta}_{\ell}, \hat{\vect{h}}_{\ell}, \hat{\nu}_{\ell}, \hat{\vect{K}}_{\ell} \}$ with $ \{\hat{\eta}_{i, \ell}\}_{i=1,\ell=1}^{n, \ \ \ m}$ \\
			
		}
		\UNTIL{ convergence}
	\end{algorithmic}
	\label{alg:VBEM}
\end{algorithm}

In the VB-M step, we compute the joint distribution $q( \mathcal{K}, \mathcal{M}, \mathcal{S} )$
from the responsibilities $ \{\hat{\eta}_{i, \ell}\}_{i=1,\ell=1}^{n, \ \ \ m}$ and the weights $\{ w_i \}^{n}_{i=1}$.
In our framework, the weight $\{ w_i \}^{n}_{i=1}$ can be computed as $w_i = \tilde{W}(\vect{s}_i, \vect{\tau}_i)$. 
The joint distribution $q( \mathcal{K}, \mathcal{M}, \mathcal{S} )$ is given by
\begingroup\makeatletter\def\f@size{8.5}\check@mathfonts
\def\maketag@@@#1{\hbox{\m@th\large\normalfont#1}}
\begin{align}
q( \mathcal{K}, \mathcal{M}, \mathcal{S} )  =  \textrm{Dir}(\mathcal{K}|\hat{\alpha})
 \prod_{\ell=1}^{m} 
\mathcal{N}( \vect{\mu}_{\ell} | \hat{\vect{h}}_{\ell}, ( \hat{\beta}_{\ell} \vect{S}_{\ell} )^{-1} ) \mathcal{W}( \vect{S}_{\ell} | \hat{K}_{\ell}, \hat{\nu}_{\ell} ),
\label{eq:VB-M}
\end{align}
\endgroup
where  
\begingroup\makeatletter\def\f@size{9}\check@mathfonts
\def\maketag@@@#1{\hbox{\m@th\large\normalfont#1}}
\begin{equation}
\hat{\gamma}_{\ell}  = \sum_{i=1}^{n} w_i \hat{\eta}_{i,\ell} , \ \ 
\hat{\vect{c}}_{\ell} = \frac{1}{\hat{\gamma}_{\ell}} \sum_{i=1}^{n} w_i \hat{\eta}_{i,\ell} \vect{x}_{i}, \ \
\hat{\vect{h}_{\ell}} = \frac{\hat{\gamma_{\ell}}}{\hat{\beta_{\ell}}} \hat{\vect{c}}_{\ell}, \nonumber
\end{equation}
\endgroup
\begin{equation}
\hat{\alpha}_{\ell}  = \alpha_{0} + \hat{\gamma_{\ell}}, \ \
\hat{\beta_{\ell}} = \beta_{0} + \hat{\gamma_{\ell}}, \ \
 \hat{\nu_{\ell}} = \nu_0 +  \hat{\gamma_{\ell}}, \nonumber
\end{equation}
{\scriptsize
\begin{equation}
\hat{\vect{K}}_{\ell}  = \left( \vect{K}_0^{-1} + \sum_{i=1}^{n} w_i \hat{\eta}_{i, \ell} ( \vect{x}_{i} - \hat{\vect{c}}_{\ell} )( \vect{x}_{i} - \hat{\vect{c}}_{\ell} )^{\top} + \frac{\beta_0\hat{\gamma}_{\ell}}{\beta_0 + \hat{\gamma}_{\ell}} \hat{\vect{c}}_{\ell}\hat{\vect{c}}_{\ell}^{\top} \right)^{-1}, \nonumber
\end{equation}	
}

\noindent and $\mathcal{W}( \vect{S} | \vect{K}, \nu )$ is the Wishart density with $\nu$ degrees of freedom.
$\alpha_{0}$, $\beta_{0}$, $\nu_0$, and $\vect{K}_{0}$ are hyperparameters.

\end{document}